\documentclass[final,1p,times]{elsarticle}

\usepackage{amssymb}

\usepackage{lineno}

\usepackage{svg}
\usepackage[hyphens]{url}
\usepackage{parskip}
\usepackage{multicol}
\usepackage{hyperref}
\usepackage{subfig}
\usepackage{array}
\usepackage{algorithm, algorithmic}
\usepackage{gensymb}
\usepackage{comment}

\usepackage{amsmath}

\usepackage{glossaries}
\newacronym{ev}{EV}{Electric Vehicles}
\newacronym{ems}{EMS}{Energy Management System}
\newacronym{bems}{BEMS}{Building-Energy Management System}
\newacronym{dhw}{DHW}{Domestic Hot Water}
\newacronym{esu}{ESU}{Energy Storage Units}
\newacronym{soc}{SoC}{State of Charge}
\newacronym{ppo}{PPO}{Proximal Policy Optimization}
\newacronym{pcc}{PCC}{Point of Common Coupling}
\newacronym{tes}{TES}{Thermal Energy Storage}
\newacronym{der}{DER}{Distributed Energy Ressources}
\newacronym{hvac}{HVAC}{Heating Ventilation and Air-Conditioning}
\newacronym{pv}{PV}{Photovoltaic}
\newacronym{flc}{FLC}{Fuzzy Locig Control}
\newacronym{pso}{PSO}{Particle Swarm Optimization}
\newacronym{mpc}{MPC}{Model Predictive Control}
\newacronym{rl}{RL}{Reinforcement Learning}
\newacronym{ann}{ANN}{Artificial Neural Network}
\newacronym{mdp}{MDP}{Markov Decision Process}

\usepackage{framed}
\usepackage{nomencl}
\makenomenclature

\usepackage{ifthen}
\renewcommand{\nomgroup}[1]{%
\ifthenelse{\equal{#1}{E}}{\item[\textbf{Energy systems}]}{%
\ifthenelse{\equal{#1}{R}}{\item[\textbf{RL modelling}]}{}}
}

\renewcommand*\nompreamble{\begin{multicols}{2}}
\renewcommand*\nompostamble{\end{multicols}}

\journal{Energy Reports}

\begin{document}

\begin{frontmatter}



\title{Data-Driven Policy Mapping for Safe RL-based Energy Management Systems\footnote{published in Energy reports journal : Volume 13, June 2025, Pages 1888-1909.}}


\author[label1]{Théo Zangato}
\author[label2]{Aomar Osmani}
\author[label1]{Pegah Alizadeh}

\affiliation[label1]{organization={LIPN, CNRS-UMR 7030, Université Sorbonne Paris Nord},
            city={Paris},
            country={France}}
\affiliation[label2]{organization={INSA Rouen, LITIS, Université de Normandie},
            city={Rouen},
            country={France}}

\begin{abstract}
    Increasing global energy demand and renewable integration complexity have placed buildings at the center of sustainable energy management. We present a three-step reinforcement learning (RL)-based Building Energy Management System (BEMS) that combines clustering, forecasting, and constrained policy learning to address scalability, adaptability, and safety challenges.
    First, we cluster non-shiftable load profiles to identify common consumption patterns, enabling policy generalization and transfer without retraining for each new building. Next, we integrate an LSTM-based forecasting module to anticipate future states, improving the RL agent’s responsiveness to dynamic conditions. Lastly, domain-informed action masking ensures safe exploration and operation, preventing harmful decisions.
    Evaluated on real-world data, our approach reduces operating costs by up to 15\% for certain building types, maintains stable environmental performance, and quickly classifies and optimizes new buildings with limited data. It also adapts to stochastic tariff changes without retraining. Overall, this framework delivers scalable, robust, and cost-effective building energy management.
\end{abstract}




\end{frontmatter}


\begin{table*}[!t]   
\begin{framed}
\nomenclature[E]{$COP$}{Coefficient of Performance}
\nomenclature[E]{$\mathcal{Q}$}{Condensor capacity ($J$)}
\nomenclature[E]{$\dot{W}$}{Instantaneous compressor power ($W$)}
\nomenclature[E]{$\dot{m}$}{Mass flow rate ($kg.s^{-1}$)}
\nomenclature[E]{$C_p$}{Specific heat ($J.kg^{-1}.K^{-1}$)}
\nomenclature[E]{$T$}{Temperature ($K$ or $\degree C$)}
\nomenclature[E]{$\eta$}{Efficiency}
\nomenclature[E]{$h$}{Enthalpy of refrigeant ($KJ.kg^{-1}$)}

\nomenclature[E]{$I$}{Current ($A$)}
\nomenclature[E]{$V$}{PV array output voltage ($V$)}
\nomenclature[E]{$N_s$}{Number of cells connected in series}
\nomenclature[E]{$N_p$}{Number of modules connected in parallel}
\nomenclature[E]{$q$}{Charge of an electron}
\nomenclature[E]{$k$}{Boltzmann's constant}
\nomenclature[E]{$\alpha$}{Cell deviation from ideal p-n junction characteristics}
\nomenclature[E]{$E$}{Energy ($J$ or $W$)}
\nomenclature[E]{$S$}{Solar irradiation( $mW/cm^2$)}

\nomenclature[E]{$P$}{Power ($W$)}
\nomenclature[E]{$v$}{Speed ($m.s^{-1}$)}
\nomenclature[E]{$\mathcal{A}$}{Area ($m^2$)}
\nomenclature[E]{$\rho$}{Density ($kg/m^3$)}
\nomenclature[E]{$C_{p_{\text{wind}}}$}{Rotor power coefficient}
\nomenclature[E]{$F$}{Correction factor}
\nomenclature[E]{$a_{\text{ang}}$}{Angular induction factor}
\nomenclature[E]{$a_{\text{axial}}$}{Axial induction factor}
\nomenclature[E]{$\lambda$}{Local Tip Speed ration (TSR)}

\nomenclature[E]{$SoC$}{State of Charge (\%)}
\nomenclature[E]{$C_n$}{Battery nominal capacity ($mAh$)}

\nomenclature[E]{$Q$}{Thermal energy ($J$)}

\nomenclature[R]{$L$}{Non-shiftable loads ($kWh$)}
\nomenclature[R]{$E^{th}$}{Electrical consumption associated with thermal needs ($kWh$)}
\nomenclature[R]{$E^{ESU}$}{Energy transfers in storage units ($kWh$)}
\nomenclature[R]{$E^{pv}$}{Energy produced by renewable sources ($kWh$)}
\nomenclature[R]{$H$}{State of Charge}
\nomenclature[R]{$E^r$}{Energy acquired from the grid}
\nomenclature[R]{$C$}{Cost of a kWh of energy (\$)}
\nomenclature[R]{$K$}{Set of temporal features}
\nomenclature[R]{$S$}{Set of states}
\nomenclature[R]{$A$}{Set of actions}
\nomenclature[R]{$P$}{Probability transition}
\nomenclature[R]{$R$}{Reward function}
\nomenclature[R]{$\gamma$}{Discount factor}
\nomenclature[R]{$\pi$}{Policy}
\nomenclature[R]{$\tau$}{Trajectory}
\nomenclature[R]{$\mathcal{T}$}{Maximum episode length}
\nomenclature[R]{$\beta$}{Action space boundaries}

\printnomenclature
\end{framed}
\end{table*}

\section{Introduction}
\label{sec:intro}

The world's global electricity energy consumption is constantly growing \cite{iea2021elec}. The growing use of electrical and connected devices used in our daily lives and the electrification of personal transportation with recent integration of \acrfull{ev} suggest that global energy demand will grow by more than a quarter to 2040, and will continue to grow for decades to come \cite{garcia2019energy}.
In this regard, electric power systems, which are central parts in the industrialization and development of territories, are at the core of the transformation challenges of our society, as they transition to smart grids to become more sustainable, distributed, dynamic, intelligent and open \cite{chen2022reinforcement}. An important factor of the explosion in consumption is the growing energy demand of buildings, especially in households, as they represent almost a third of the total consumption and an equivalent share in greenhouse gas emissions \cite{iea2022buildings}. Several elements contribute to rising energy demand in buildings such as the  the adoption of electric heat pumps for heating and cooling, rising standards of living, and the generalization of \acrshort{ev}s \cite{iea2021ev}.\\
Additionally, renewable energy sources are increasingly significant in global energy production, comprising nearly 30\% of the global energy generation. Buildings directly contribute to their integration as distributed \acrfull{pv} installations represent 40\% of total PV installations \cite{iea2022pvhousehold}. These trends highlight the central role of households and their structures in global energy consumption and generation. However, the integration of intermittent renewable energy sources, combined with the large-scale regional interconnection of energy systems, has transformed sustainable energy and electric systems into highly dynamic and volatile large-scale entities. As a result, the dynamic of supply and demand is projected to drive up electricity prices, strengthening the financial burden on households, as energy represents up to 25\% of European household expenditures in 2021 according to Eurostat \footnote{source: \url{http://data.europa.eu/88u/dataset/e3td1ejcprfbhotlntxwa}}.

Therefore, buildings and individual residences occupy a key role in reshaping the current energy production and consumption paradigm in regard of the consequent share of energy consumption to which their are associated, and their impact on the penetration of \acrfull{der}. Moreover, optimizing energy consumption and tackling emissions generated by buildings is key as the International Energy Agency foresees a 20\% growth in floor area by 2030 with the rise and urbanization of developing economies \cite{iea2022buildings}. As those challenges emerge, the need for innovative solutions and adaptive strategies addressing building's consumption and energy production becomes essential to ensure stability, efficiency, and sustainability \cite{HEIDARY2023109105}. In response, \acrfull{ems}  have emerged as vital tools in the energy landscape. They are designed to monitor, control, and optimize energy production and consumption within specific systems. Given the specific needs of buildings, the concept of \acrfull{bems} has gained traction. With the escalating complexity of energy networks and the growing integration of renewable energy sources, they become important in ensuring efficient and consistent energy use, while maximizing the reliance on renewable sources and counteracting their intermittent nature \cite{chaudhary2021review}. By offering real-time insights and automation capabilities, \acrshort{ems} not only reduces energy wastage and costs but also bolsters reliability, mitigates environmental impact, safeguards components from potential damage due to overloading, and reduces end-user energy bills \cite{olatomiwa2016energy}.

Different optimization solutions were proposed for \gls{ems} including approaches that rely on mathematical or expert models, including Model Predictive Control (MPC) \cite{vasquez2023balancing} and Ruled-Based Control (RBC) \cite{10087163}. However those approach suffer from extensive computational requirements, reliance on expert knowledge or precise sensors which prevent their use in high dimensional environments. As a result, other solutions like Fuzzy Logic Control (FLC) \cite{rodriguez2024simple} or Particle Swarm Optimization (PSO) \cite{faria2023optimal} were also investigated.\\
Reinforcement Learning is the main machine learning paradigm for sequential decision making. Consequently, RL-based \gls{ems} has emerged as an active research area \cite{GANESH2022111833, VENKATASATISH202227646, DBLP:journals/iotj/YuQZSJG21, DBLP:journals/corr/abs-2008-05074} due to notable advancements, particularly in the gaming field \cite{DBLP:conf/aaai/LampleC17, DBLP:journals/corr/MnihKSGAWR13, DBLP:journals/corr/abs-1712-01815}. RL is a machine learning approach that learns a sequential decision-making policy by an agent that directly interacts with its environment \cite{DBLP:books/lib/SuttonB98}. It distinguishes itself as a suitable approach for energy management systems by its ability to focus on optimizing long-term rewards, that consider future impacts. It can balance multiple objectives, as well as navigate complex, uncertain and high dimensional environments under constraints. RL's adaptability and capability to approximate hard-to-measure phenomena makes it a strong choice for complex systems. Nonetheless, it faces challenges with generalization and sample efficiency, limiting the ability to apply and transfer a learned model to other instances with varying data distribution, without retraining a whole new model \cite{DBLP:conf/aaai/ZhangKOB21, pmlr-v97-cobbe19a}.

Using the sequential nature of energy consumption to improve \acrshort{ems} algorithms is a well-established practice. Time-series analysis have been combined with various methods as a way to feed existing knowledge into the given control method. Specifically, the majority of research has focused on time-series forecasting, given its comprehensive documentation \cite{DEB2017902}. However, the use of time-series data for generalization across different scenarios remains underexplored \cite{manfren2022data, hussain2024energy}.

This paper aims to achieve two primary objectives: firstly, to introduce an effective and safe method for managing the charging cycles of electrical or thermal storage systems in buildings; and secondly, to propose a policy generalization approach that harnesses knowledge extracted from time-series data.

We propose a reinforcement learning based \acrshort{bems} while addressing sample inefficiency concerns and generalization to unknown building consumption patterns. Our solution involves leveraging prior knowledge into the learning model, allowing agents to focus on task optimization rather than learning environmental constraints from scratch. Moreover, we develop a specific subgroup generalization method, reducing the need to adapt to a broad range of buildings based on varying energy needs.
These two key aspects of our solution ensures controlled and safe exploration during training, boosting agent effectiveness in real-world scenarios while minimizing extensive modeling and associated approximations \cite{BEAUDIN2015318}. Our method includes analyzing historical energy consumption data to create typical profiles by clustering load curves from buildings and to train an LSTM based prediction module for real-time environmental observations. On the other hand, specific domain knowledge informs decision-making probabilities, and during inference, a classification module assesses load profile similarities, facilitating policy generalization to similar buildings. Our approach enables the buildings to limit their energy expenses as well as network operators by prioritizing stored energy in peak period consumption. We demonstrate our method using the Citylearn environment, using a set of real buildings comprising different consumption patterns, energy devices and power generation \cite{vazquez2019citylearn}.\\
Key contributions can be summarized as below:
\begin{itemize}
    \item Develop a safe RL-based Building Energy Management System (BEMS)
    \item Integrate prior and learned domain knowledge into the learning model
    \item Propose a policy generalization approach leveraging time-series data
    \item Demonstrate its effectiveness in real-world scenarios
\end{itemize}
The structure of our work is outlined as follows: in Section \ref{sec:bems}, we present the context of \acrshort{bems} and modelling avenues of the principal components. Section \ref{sec:liter} offers a comprehensive overview of the methodologies and strategies existing in the literature. In Section \ref{sec:problem-formulation} we formulate our \acrshort{bems} problem and propose a mathematical model for the principal components while Section \ref{sec:approach} is dedicated to detailing our approach, where we implement a data-driven policy mapping method for Reinforcement Learning-based \acrshort{ems} agents. Finally we discuss our approach and conclude in Section \ref{section:conclusion}. 

\section{BEMS Formulation}
\label{sec:bems}

\begin{figure}
    \centering
    \includegraphics[scale=0.28]{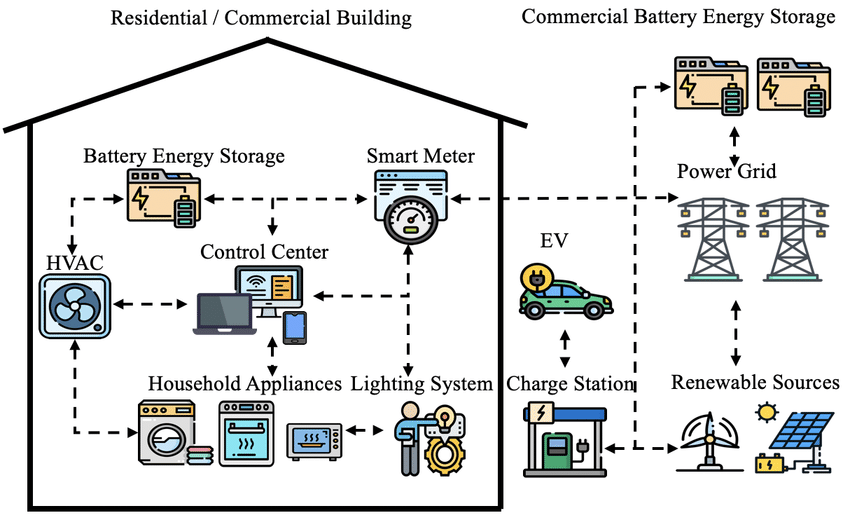}
    \caption{Example of a BEMS in a commercial/residential building \cite{zhang2021data}.}
    \label{fig:bems}
    
\end{figure}

This section provides a clear definition of a \acrshort{bems} and a detailed modelling of its main components, which are recurrent in the literature. Each component has its own characteristics and interacts in a specific way with all or part of the system. The list of components presented below is not exhaustive, but groups together the most important concepts, and models. Figure \ref{fig:bems} shows a typical \acrshort{bems} architecture.

\subsection{Ecosystem}
\paragraph{\textbf{Energy management}}
Energy management aims at consistently and efficiently monitor and manage the use of energy, with a specific focus on optimizing energy costs based on factors such as user characteristics, financial capabilities, energy demands, available funding opportunities, occupant comfort, or emission reductions. In both residential and non-residential buildings, systems and strategies that embody this idea can be defined as \acrfull{bems} \cite{mariano2021review}. In this paper, we specifically concentrate on active methods that involve real-time management of a building's appliances, utilizing load signals, sensors, and historical data.

\paragraph{\textbf{Smart grid}}
Smart grids are advanced electrical grids that monitor and manage real-time electricity distribution from a decentralized power generation scheme. They use innovative technologies to coordinate power generation from diverse sources and align with user demand. This approach is essential for adapting to a changing power supply systems that now include of grid-connected and off-grid systems and increased volatility in demand along the day, with peaks typically in midday and evening \cite{zubi2018lithium}.

Furthermore, smart grids are crucial for \acrshort{bems}, operating either within the entire grid or a subsection, such as a micro-grid. Unlike traditional systems focused on individual buildings, smart grids aim to optimize energy use and costs at the network level, integrating objectives like reducing energy consumption, managing costs, and maintaining network safety and reliability by addressing challenges like consumption peaks and load variations amongst other \cite{iea2022smartgrid}. Each building within this network is considered a component of the broader management system, highlighting a shift towards comprehensive network management rather than focusing solely on end-user optimization.

\paragraph{\textbf{Smart building}}
Just like smart grids, smart building integrate a share of connected energy devices that can be leveraged in \acrshort{ems} as to optimize their energy consumption. It can integrate personal energy generation systems and either be connected to the common grid or completely off-grid. The smart building is the most common area of operation in the \acrshort{bems} field as they enable to better integrate users and their production capacities in the global energy mix.
Generally, objectives that only include the building being treated are considered, whether it is connected to the distribution network or alone. In modelling at the scale of a single building, it is easier to model a larger set of components, without risking an explosion in the number of parameters, than in the case of micro-grids.

\subsection{Energy consumption systems}
When it comes to intelligent buildings, thermal equipment plays an important role as they represent nearly 80\% of households energy consumption \cite{iea2022buildings}. Different types of thermal appliances exists such as Heat Pumps, air-conditioning systems or electric heaters, all of which can be controlled by an EMS in a smart building and are often summarized under the \acrfull{hvac} designation.

More generally, energy consumption can be categorized into two distinct groups: 1) shiftable loads and 2) non-shiftable loads. Shiftable loads refer to activities or devices for which their use can be deferred in time. The user or system can choose to voluntarily adjust its operation time without any degradation. They typically can be delayed to times where \textit{Time-of-Use} tariffs or grid consumption is low. This includes devices such as \acrshort{ev}s, ovens or washing machines and industrial processes. On the other hand, non shiftable loads represents devices or activities which their operating schedules cannot be adjusted such as essential household appliances like refrigeration systems or fixed production schedules manufacturing processes.

\paragraph{\textbf{Heat Pump}}
Heat pumps play a crucial role in energy transition programs. A key efficiency advantage lies in the fact that heat pumps do not generate heat; instead, they extract heat from a source, amplifying it, then transfer it to the desired environment. This transfer mechanism requires significantly less energy compared to traditional heating techniques. Several work have studied the optimization of heat-pumps in buildings. Specifically, heat-pumps systems coupled with solar systems. In \cite{cao2020thermal}, authors optimized a Direct-Expansion Solar-Assisted Heat Pump Water Heater (DX-SAHPWH) using NSGA-II, targeting the coefficient of performance (COP) and solar collector efficiency (SCE). Bi-criteria optimization improved COP by 20\% with a 1.6\% SCE reduction, outperforming single-objective methods. Decision-making techniques (LINMAP, TOPSIS, Shannon’s entropy) identified the optimal configuration, showcasing advanced genetic algorithm applications in heat pump optimization.\\
Other techniques such as a data-driven Model Predictive Control (MPC) framework were used for optimizing heat pump systems in supplying district heating. Using machine learning models for heating load and indoor state prediction, coupled with Particle Swarm Optimization (PSO), \cite{wei2022data} minimized energy consumption while maintaining indoor comfort. Their framework achieved energy savings of up to 12.37\% in medium-load scenarios and was validated through both offline simulations and onsite tests, demonstrating a 12.19\% energy-saving ratio under real-world conditions. One can find a specific review of Reinforcement Learning applications to control of \acrshort{hvac} systems in the work of Sierla et al \cite{en15103526}.

Heat pumps, serving dual purposes for both cooling and heating, are set to become the default devices for climate control in net-zero scenarios. As such, the focus will be exclusively on the method for evaluating their performance. The Coefficient of Performance (COP) is commonly used to evaluate the unit performance of such systems, it represents the ratio of the useful thermal power produced divided by the mechanical power used to generate this energy. It can be described as follows \cite{byrne2022research, mohanraj2018research}:
\begin{equation}
    \label{eq:cop-heat-pump}
    COP = \frac{\mathcal{Q}}{\dot{W}}
\end{equation}
with $\dot{W}$ the instantaneous compressor power (W) consumption given by:
\begin{equation}
    \dot{W} = \frac{\dot{m_r}X}{\eta_{mech}\times \eta_{elec}}
\end{equation}
with $X= (h_3 - h_4)$ for heating and $X=(h_2 - h_1)$ for cooling, $\eta_{mech}$ the compressor mechanical efficiency, $\eta_{elec}$ the compressor electrical efficiency and $h$ the enthalpy of the refrigerant at different steps of the thermodynamic cycle.

The condensor capacity $\mathcal{Q}$ is defined as:
\begin{equation}
    \mathcal{Q} = \dot{m}.C_p.\Delta T
\end{equation}
where $\dot{m}$ ($kg.s^{-1}$) is the mass flow rate, $C_p$ the specific heat ($J.kg^{-1}.K^{-1}$) and $\Delta T$ the discrepancy of temperature.

\subsection{Energy production systems}

Energy production in smart buildings differs on the use case. The building can be connected to the power network that can supply an unlimited amount of energy at a certain cost. The energy provided has a particular cost that depends on \textit{Time-Of-Use} tariffs and an associated carbon emission costs. Production can also be assured by renewable sources as they represent a cost-free, emission-free solution for energy production. They usually need to be coupled with energy storage systems.
It is also a central feature of off-grid systems that rely exclusively on these options for energy production. Photovoltaic panels are the most common in the literature as well as in terms of total installed power, but solar water heaters and wind turbines are also used.

\paragraph{\textbf{\acrfull{pv}}}
Solar PV generation reached almost 1300 $TWh$ in 2022 due to its modular technology that can be deployed from small residential roof-top systems up to utility-scale power generation installations. Their integration, utilizing energy directly or through storage systems, reduces grid load and energy costs \cite{iea2022pvhousehold}.\\
Photovoltaic systems, due to their scalability and ease of deployment, have gained significant research interest, particularly in optimizing their integration and the utilization of generated energy. In \cite{9113746}, authors propose a Multi-Agent RL framework with an integrated attention model for voltage regulation in distribution systems with high photovoltaic penetration. By leveraging local observations, the method trains decentralized agents to develop coordinated control strategies that optimize reactive power dispatch of PV inverters. The approach achieves up to 95.1\% of the centralized control's performance while reducing communication infrastructure costs. Similarly, the work in \cite{godina2018optimal} investigates the integration of \acrshort{pv} microgeneration with a Model Predictive Control (MPC) framework for residential HVAC optimization under varying Time-of-Use rates. A two-level control signal and weight tuning were applied to minimize energy costs. The method reduced costs by 75\% with PV integration compared to non-PV scenarios, demonstrating the effectiveness of combining MPC and PV systems for energy-efficient residential management. For a broader perspective, Gaviria et al. provide a comprehensive review of machine learning techniques applied in PV systems, offering further insights into advancements in the field \cite{gaviria2022machine}.

\begin{figure*}[htp!]
\centering
\include[inkscapelatex=false,scale=0.9]{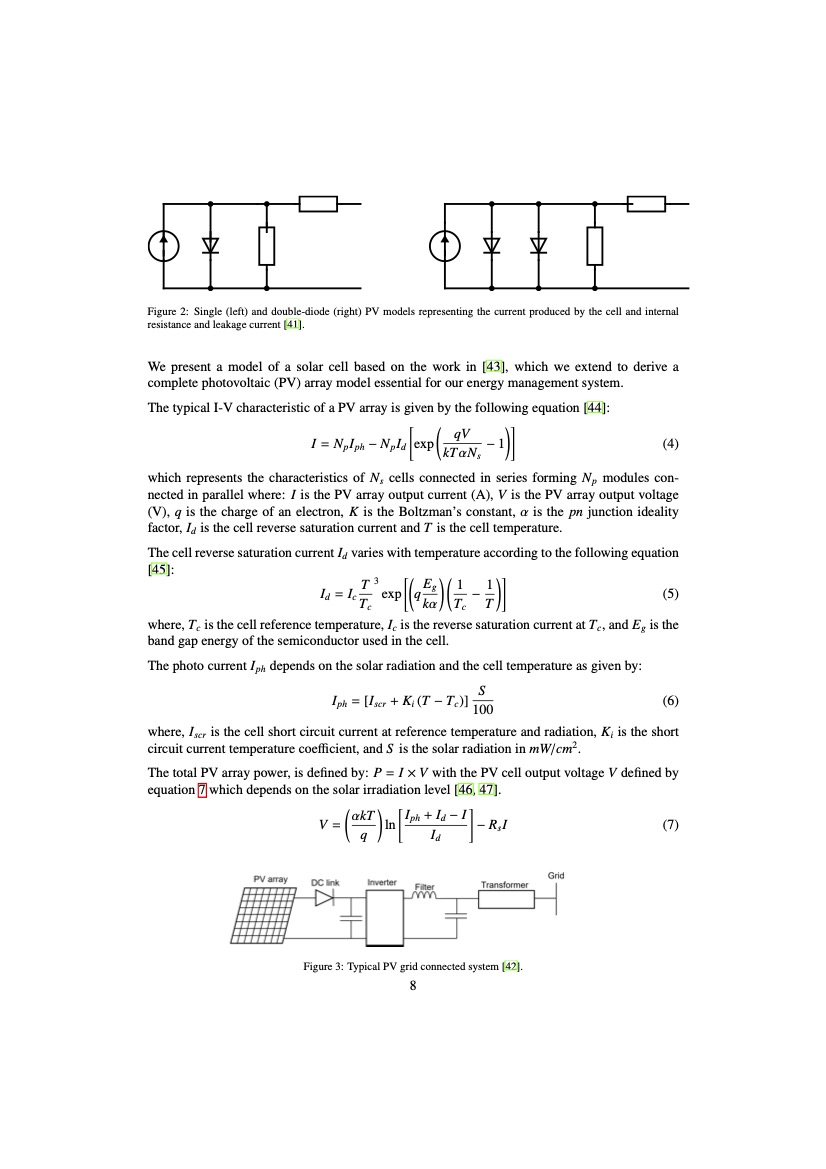}
   \caption{Single (left) and double-diode (right) PV models representing the current produced by the cell and internal resistance and leakage current \cite{HASAN201675}.}
    \label{fig:pv-model}
    
\end{figure*}

The operating principle of solar panels is based on the photovoltaic effect, which transforms the sun's rays into electricity using a semiconductor device known as the Photovoltaic cell. Figure \ref{fig:pv-model} presents two common electrical diode models of solar cells.

\begin{figure}[b]
    \centering
    \includegraphics[scale=0.45]{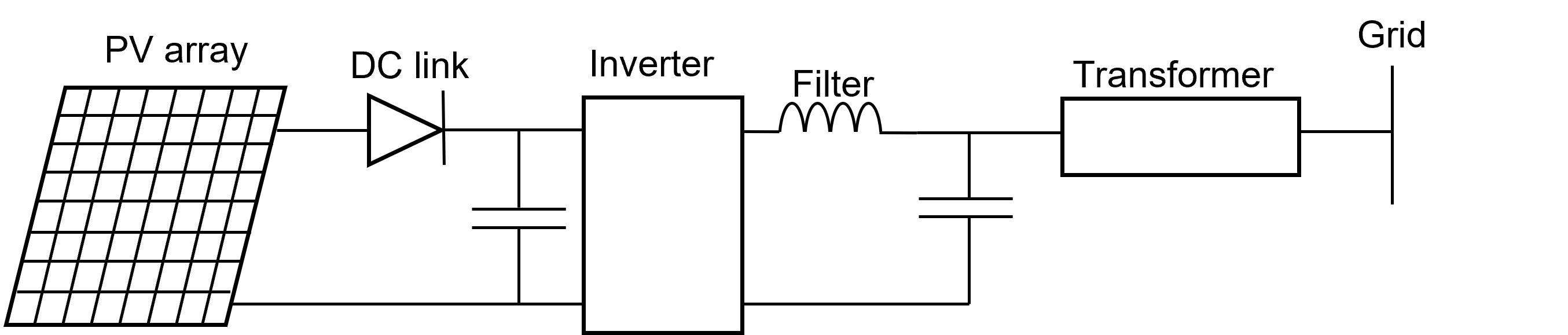}
    \caption{Typical PV grid connected system \cite{abdul2014}.}
    \label{fig:pv-system}
    
\end{figure}

PV arrays are a concatenation of multiple cells that produce DC output current energy converted into AC and injected into the grid through an inverter. Energy production is highly nonlinear and dependent on exterior factors such as the sun’s geometric location, solar irradiance and ambient temperature. Figure \ref{fig:pv-system} represents a classic PV grid connected system.

We present a model of a solar cell based on the work in \cite{singh2013solar}, which we extend to derive a complete photovoltaic (PV) array model essential for our energy management system.

The typical I-V characteristic of a PV array is given by the following equation \cite{hussein1995maximum}:
\begin{equation}
    \label{eq:pv-iv}
    I = N_pI_{ph} - N_pI_d \left[\exp \left(\frac{qV}{kT\alpha N_s}-1 \right)\right]
\end{equation}
which represents the characteristics of $N_s$ cells connected in series forming $N_p$ modules connected in parallel where: $I$ is the PV array output current (A), $V$ is the PV array output voltage (V), $q$ is the charge of an electron, $K$ is the Boltzman’s constant, $\alpha$ is the $pn$ junction ideality factor, $I_d$ is the cell reverse saturation current and $T$ is the cell temperature.

The cell reverse saturation current $I_d$ varies with temperature
according to the following equation \cite{vachtsevanos1987hybrid}:
\begin{equation}
    \label{eq:pv-reverse-saturation-current}
    I_d = I_c \frac{T}{T_c}^3 \exp \left[\left(q\frac{E_g}{k\alpha}\right)\left(\frac{1}{T_c}-\frac{1}{T} \right) \right]
\end{equation}
where, $T_c$ is the cell reference temperature, $I_c$ is the reverse saturation current at $T_c$, and $E_g$ is the band gap energy of the semiconductor used in the cell.

The photo current $I_{ph}$ depends on the solar radiation and the cell temperature as given by:
\begin{equation}
    \label{eq:pv-photo-current}
    I_{ph} = \left[ I_{scr}+K_i \left( T - T_c\right)\right] \frac{S}{100}
\end{equation}
where, $I_{scr}$ is the cell short circuit current at reference temperature and radiation, $K_i$ is the short circuit current temperature coefficient, and $S$ is the solar radiation in $mW/cm^2$.

The total PV array power, is defined by: $P = I \times V$
with the PV cell output voltage $V$ defined by equation \ref{eq:pv-output-voltage} which depends on the solar irradiation level \cite{hua1998implementation, altas2007photovoltaic}.
\begin{equation}
    \label{eq:pv-output-voltage}
    V = \left( \frac{\alpha kT}{q} \right) \ln\left[ \frac{I_{ph}+I_d-I}{I_d} \right] - R_sI
\end{equation}

\paragraph{\textbf{Wind Turbine}}

The most productive non-hydro renewable technology is wind energy, which produces more electricity than all other comparable technologies combined with 2100 TWh of electricity in 2022 \cite{iea2023wind}. 
Wind turbine optimization include several aspects, including the adjustment of turbine parameters and the strategic placement of turbines within wind farms.
For example, in \cite{emami2010new} authors introduces a genetic algorithm approach for optimizing wind turbine placement within wind farms, incorporating a new coding scheme and a multi-objective function with adjustable cost and efficiency weights. Unlike prior methods, this approach eliminates the need for subpopulations and significantly improves computation efficiency. Applied to three wind scenarios—uniform, multi-directional, and variable wind—the method achieved higher power outputs, greater efficiency, and reduced turbine counts compared to earlier works. In contrast, \cite{fernandez2022actor} proposes an Actor-Critic Reinforcement Learning framework for optimizing Variable-Speed Wind Turbine controllers by jointly adjusting blade pitch and generator torque parameters. Unlike traditional decomposed approaches, this method uses an augmented input space to fine-tune parameters for varying wind conditions. Their model achieved up to a 19.12\% reduction in average power error compared to baseline controllers. A complete review of the optimization techniques and strategies applied to wind turbine performance optimization can be found in the work of Chehouri et al. \cite{chehouri2015review}.

A wind turbine is an electromechanical energy conversion device. It captures kinetic energy from wind motion to rotate a turbine rotor consisting of blades that converts the wind energy into a rotational energy. The dynamic interaction between the wind and the response of the turbine determines the amount of kinetic energy that can be extracted. The resulting energy is then transferred to the \gls{pcc} to be instantly used or stored by the building.
There is a wide range of modeling representation of turbines and especially their generator in the literature \cite{petru2002modeling}. We present the most common two types of wind turbines and an associated model \cite{neammanee2007development, santoso2007fundamental}.

The two main types of wind turbines are fixed- (FSWT) and variational-speed turbines (VSWT). The former uses a rotor that has a constant speed while the later can operate at different speeds depending on specific operation requirements such as the wind speed. Variational-speed turbines modify their rotor speed by varying the blade pitch angle and electrical generator speed. This enables the turbine to better capture wind energy but necessitate power converters to connect the turbine to the grid while fixed-speed turbines can be directly connected.

The energy efficiency and power generation of a wind-turbine depends on numerous factors \cite{santoso2007fundamental}. However the maximum theoretical power that can be harvested from the wind that activates a rotor from the rotation of the blades can be obtain from:
\begin{equation}
        \label{eq:wind-cp-ideal}
        P_{wind}(v_t) = \frac{1}{2}\rho_{\text{air}} \mathcal{A}_{turb}v^3_t
\end{equation}
where $P_{wind}$ is the power ($W$) associated with a wind speed $v_t$ ($m/s$) with $\mathcal{A}_{turb}$ the rotor area ($m^2$) and $\rho_{\text{air}}$ the air density. But in practice this quantity is reduced by a quantity $C_{p_{\text{wind}}}$, the rotor power coefficient, that measure the rotor's efficiency:
\begin{equation}
    C_{p_{\text{wind}}}(v_t) = \frac{P_{turb}(v_t)}{P_{wind}(v_t)}
\end{equation}

The maximum value of \( C_{p_{\text{wind}}} \), known as the Betz limit, is \(\frac{16}{27}\), and acts as a theoretical upper bound. However, real systems typically exhibit a value of 0.5, that can be calculated using equation \ref{eq:wind-cp} \cite{manwell2010wind}:
\begin{equation}
    \label{eq:wind-cp}
        C_{p_{\text{wind}}} = \frac{8}{\lambda^2} \int_{\lambda_h}^\lambda F \lambda^3_r a' (1-a)d\lambda_r
    \end{equation}
by assuming no drag, with $F$ the correction factor, $a_{\text{ang}}$ and $a_{\text{axial}}$ the angular and axial induction factors, $\lambda_r$ the local Tip Speed Ratio (TSR) at radius $r$ and $\lambda_h$ the local TSR at the hub.

The power delivered by a turbine is function of the wind speed crossing the blades. Wind turbines follow a typical power curve that can be approximated for VSWT as \cite{carrillo2013review}:
\begin{equation*}
    \label{eq:wt-power-curve}
P_{turb}(v_t) = 
    \begin{cases}
        0 &\text{$v_t < v_{{ci}_t}$ or $v >  v_{{co}_t}$}\\
        P_{\text{exp}}(v_t) &\text{$v_{{ci}_t} \leq v_t < v_{{r}_t}$}\\
        P_{rated} &\text{$v_{{r}_t} \leq v_t \leq v_{{co}_t}$}
    \end{cases}
\end{equation*}
where $P_{turb}$ is the electric power ($W$), $v_{ci}$ is the cut-in wind speed ($m/s$), $v_{co}$ is the cut-out wind speed ($m/s$), $v_r$ is the rated wind speed ($m/s$), $P_{rated}$ is a constant rated power ($W$) and $P_{\text{exp}}$ is the non-linear relationship between power and wind speed obtained from measurements.

\begin{figure}
    \centering
    \includegraphics[scale=0.6]{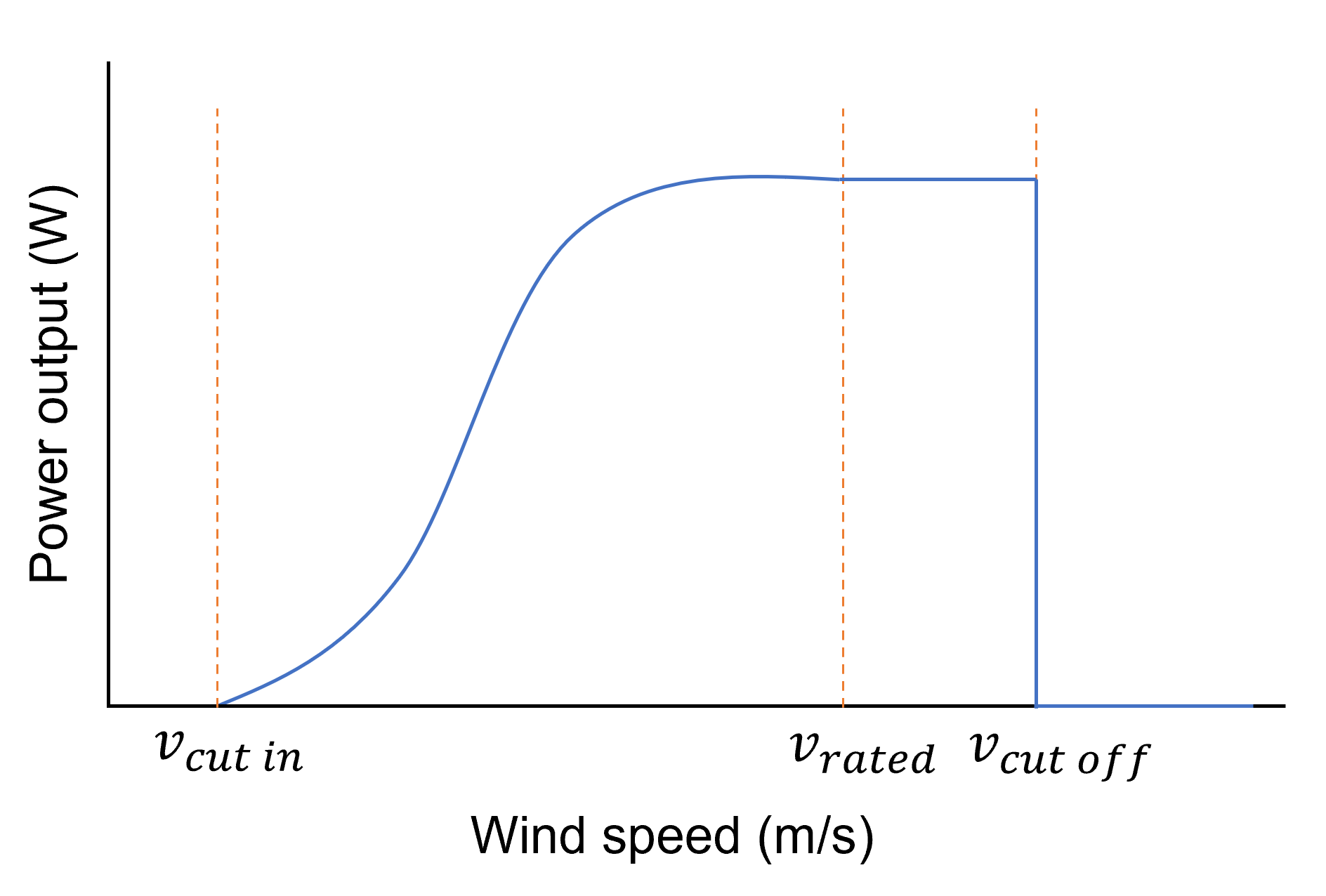}
    \caption{Representation of the wind power curve \cite{sohoni2016critical}.}
    \label{fig:wind-power-curve}
    
\end{figure}

To approximate the non-linear part $P_{\text{exp}}(v_t)$ of the wind turbine's power curve equation one can use an exponential power-curve among other methods as model using \cite{mathew2006wind}:
\begin{equation}
    P_{\text{exp}}(v_t) = \frac{1}{2}\rho_{\text{air}} \mathcal{A}_{turb}K_p \left( v^\beta_t - v^\beta_{ci_t}\right)
\end{equation}
where $K_p$ and $\beta$ are constant parameters.


\subsection{Storage Systems}
\label{sec:storage-system}
Energy storage systems or electric energy storage (EES) are not new, pumped-storage hydropower (PSH) first appeared in the 1890s. However other emerging storage systems are becoming key parts of smart grids and buildings. EES are systems that can convert electricity to another form to be released when required. This can be achieved following different techniques such as PHS, thermal and electrochemical techniques. Among the characteristics we can mention energy density, specific energy, specific power, round-trip efficiency, self-discharge rate, calendar and cycle lives, full charge and discharge times, initial cost, O\&M requirements and safety. We can divide energy storage in 2 distinct groups based on their operating scale:
\begin{itemize}
    \item \textit{Building-scale} storage encapsulate systems that can be easily integrated in buildings and meet cost and safety constraints. The most common example are \acrfull{tes} and batteries.
    \item \textit{Grid-scale} storage refers to technologies connected to the power grid that can store large amounts of energy and then supply it back directly to the grid. They can be regarded as an energy production system from a building's point of view but when the focus is shifted towards a micro-grid of several buildings, they can be regarded as the same mechanism than building-scale storage. Even if few literature explored the management of grid-scale storage they are worth mentioning as they are set to become the majority storage capacity by 2030.
\end{itemize}
In general, energy storage capacities enable greater demand response possibilities at different levels, and therefore, can relieve the network of peak consumption. In addition with short term balancing capacities, energy storage systems also provide operating reserves end ancillary services for grid stability and reliability.

\paragraph{\textbf{{Battery}}}
Batteries offer storage capacities that can be used to store the excess energy produced by renewable energies that cannot be instantly used. They can therefore delay the use of previously generated energy to be used when the consumption is at peak periods or at times when renewables don't produce any energy. They are a direct contributor to their integration. They will also become a key components in off-grid applications where solar home systems can generate up to kilowatts of energy to power basic appliances. Given its critical role in modern energy systems, battery optimization has gained strong attention of researchers.\\
For instance, \cite{liu2021operational} introduces a bi-level optimization model for the day-ahead operation of a building-level integrated energy system. This model optimizes battery charging and discharging by accounting for dynamic electricity prices and generation shortages, utilizing a Particle Swarm Optimization (PSO) algorithm. Simulation results indicate that the proposed strategy reduces total operational costs by up to 7\%.
Similarly, \cite{antoniadou2020market} presents a market-based energy management framework for building microgrids. This approach incorporates a measurement-based battery energy storage (BES) model to optimize energy dispatch under dynamic electricity pricing, while also integrating advanced lifecycle modeling to account for battery degradation. The framework demonstrates cost reductions and improved accuracy in BES performance predictions.
In another study, \cite{gao2022operational} applies deep reinforcement learning methods, specifically Deep Deterministic Policy Gradient (DDPG) and Twin Delayed Deep Deterministic Policy Gradient (TD3), to optimize off-grid operations and ensure battery safety in renewable energy systems for buildings. By employing Gaussian-based reward structures, the TD3 approach outperforms other methods, achieving a grid power purchase error below 2 kWh and ensuring battery safety over 160+ hours in a week-long simulation cycle.
For a comprehensive review of Battery Energy Storage Systems (BESS) and their optimization, including objectives, constraints, and methodologies, the work of \cite{hannan2021battery} serves as a valuable resource.

Li-ion batteries are favored for several compelling reasons. Their high energy efficiency and high energy density make them suitable choices for grid electric applications. Moreover, lithium possesses intrinsic properties that enhance its appeal for battery applications, such as high cell potential and low weight, contributing to excellent gravimetric and volumetric capacities. However, Li-ion batteries also present notable drawbacks. They are relatively expensive compared to other battery technologies and have safety concerns due to the risk of thermal runaway, which can lead to fires or explosions if not properly managed. Additionally, they experience capacity degradation over time, especially under high temperatures or improper charging cycles, which limits their lifespan. Environmental concerns related to lithium extraction and processing also pose challenges. Despite these issues, the advantages of Li-ion batteries often outweigh the disadvantages for many applications. For these reasons, this study focuses on Lithium-based batteries while considering strategies to mitigate their shortcomings.\\
Present day li-ion batteries feature a lithium-ion donator cathode, and the anode, which are respectively the positive and negative terminal of the battery. The ions move from the cathode to the anode during charging cycles and revert back during discharge. There are several types based on the type of material used that provide different properties. The main types are:
\begin{itemize}
    \item Lithium Cobalt Oxide (LCO, cathode)
    \item Lithium iron phosphate (LFP, cathode)
    \item Lithium Manganese Oxide (LMO, cathode)
    \item Lithium Titanate Oxide (LTO, anode)
    \item Lithium Nickel Cobalt Aluminium oxide (NCA, cathode)
    \item Lithium Nickel Cobalt Manganese oxide (NMC, cathode)
\end{itemize}

\begin{figure*}%
    \centering
    \subfloat[\centering Charge]{{\includegraphics[scale=0.4]{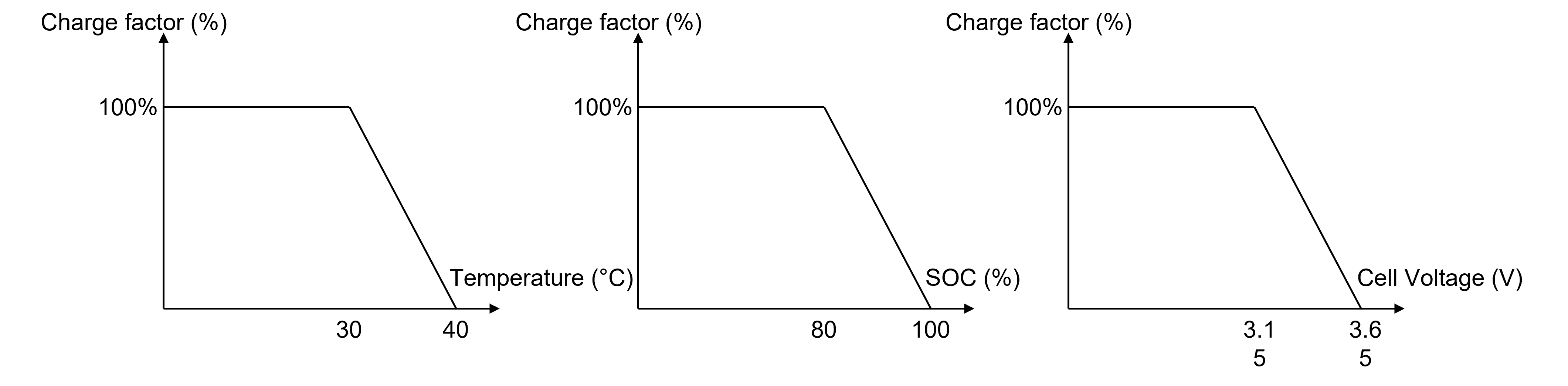} }}%
    \qquad
    \subfloat[\centering Discharge]{{\includegraphics[scale=0.4]{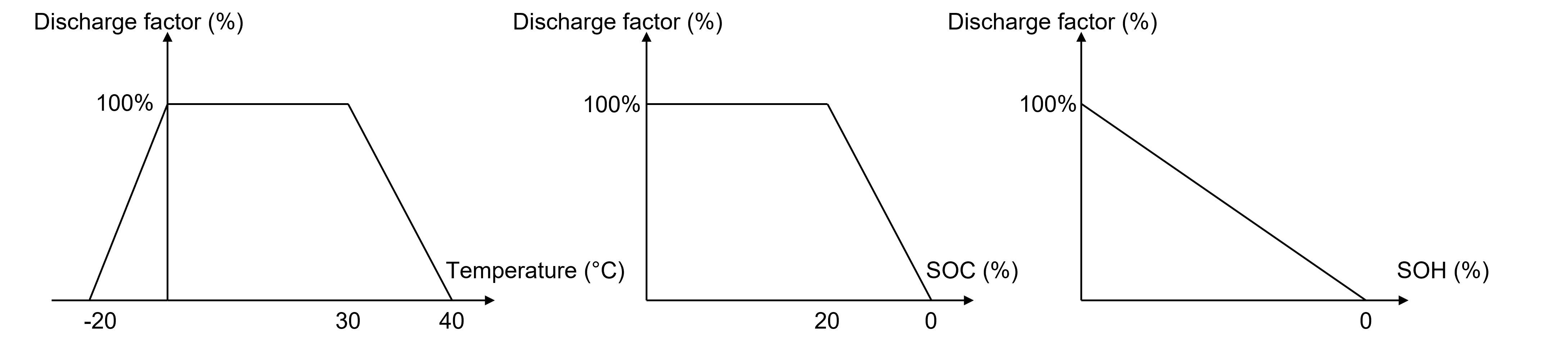} }}%
    \caption{Maximum charge and discharge current for various values of different battery parameters (Rates: 0.33C and 1C) \cite{cheng2010battery}.}%
    
    \label{fig:charge-discharge-curve-param}%
\end{figure*}

The performance of a battery when it is connected to a load or a source is based on the chemical reactions inside the battery \cite{cheng2010battery}. One of the important parameters that are required to ensure safe charging and discharging as well as to adapt the charging strategy of a battery is \acrfull{soc}. SOC is defined as the present capacity of the battery expressed in terms of its rated capacity \cite{affanni2005battery}.\\
However, measuring \acrshort{soc} is a complicated process, because it involves the measurement of a multitude of the battery's parameters such as voltage, current, temperature, as well as intrinsic information of the battery under consideration. Environmental factors and chemical reactions also complicate the estimations of the \acrshort{soc} as they bias the measurements (Figure \ref{fig:charge-discharge-curve-param}).

\begin{figure}
    \centering
    \includegraphics[scale=0.5]{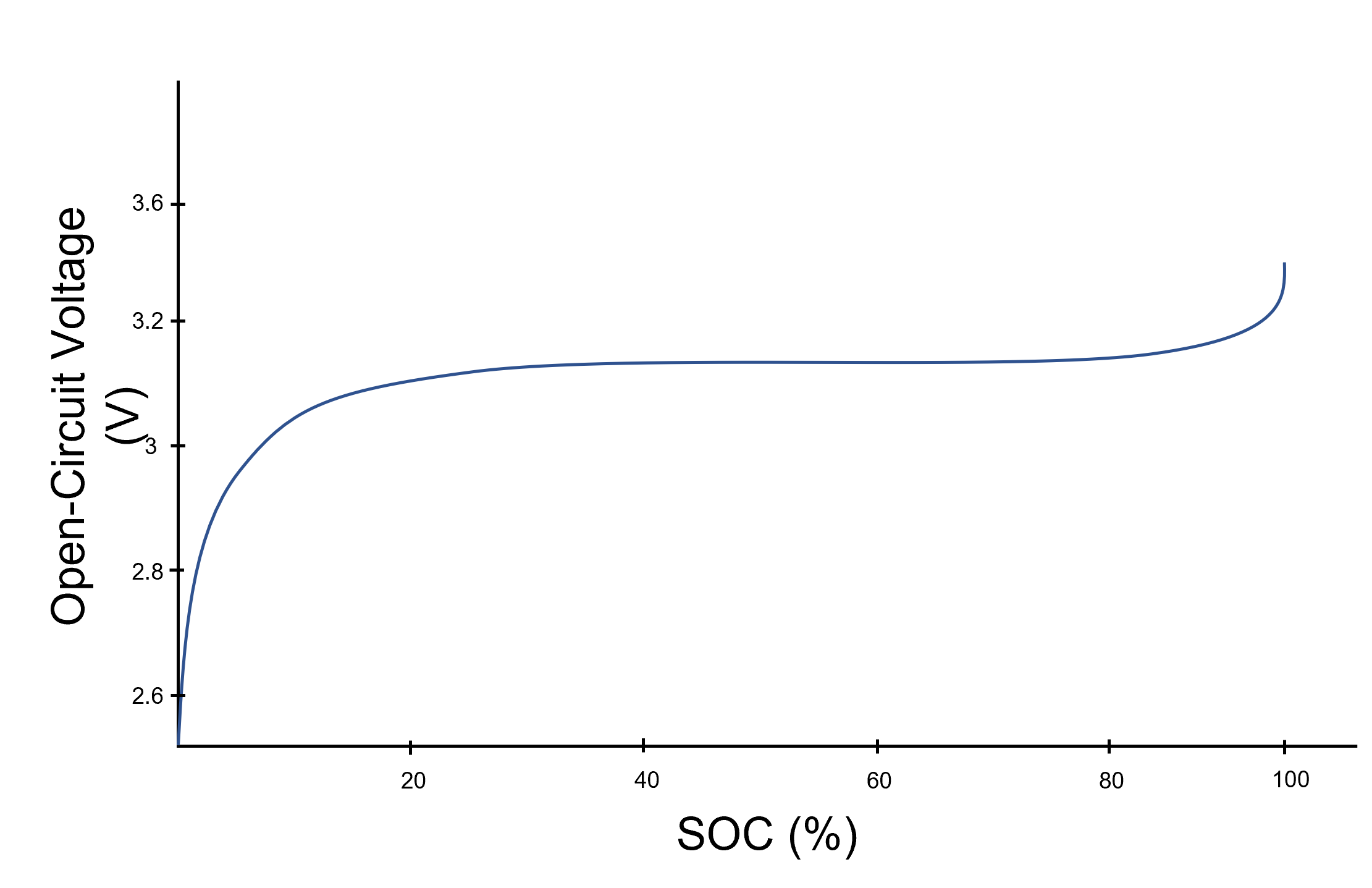}
    \caption{Typical OCV-SOC curve of a Li-ion battery.}
    \label{fig:charge-curve}
\end{figure}

Modeling battery behavior including the \acrshort{soc} is generally based on the charging and discharging curves of the studied battery. The Open Circuit Voltage-Based Estimation (OCV) method uses the OCV curve, as shown in Figure \ref{fig:charge-curve}, of a given battery to estimate the \acrshort{soc} using a look-up table with an OCV curve inversion method or a model-based method \cite{lavigne2016lithium}.
\begin{equation}
    \label{battery-ocv-estimation}
    SoC = f^{-1}(OCV)
\end{equation}
These methods suffer from time constraints and rely on the resolution of the sensors. As such, other analytical methods can be used to estimate the state of charge such as the Coulomb Counting Estimation \cite{rivera2017soc}:
\begin{itemize}
    \item Coulomb Counting Estimation (CC): It is the most common method due to its accuracy for short term estimations and is the standard method in the industry \cite{lashway2016adaptive}. It is defined by \ref{eq:battery-cc-estimation} \cite{WAAG2013548}:
    \begin{equation}
        \label{eq:battery-cc-estimation}
        SoC(t) = SoC(t_0) + \frac{1}{C_n}\int_{t_0}^{t_0+1}I_{bat}(d\tau) \times 100%
    \end{equation}
    where $SoC(t_0)$ is the initial \acrshort{soc}, $C_n$ the nominal capacity, and $I_{bat}$ is the charging/discharging current. This method, however, suffers from initial value measurement error and accumulated errors.
\end{itemize}

\paragraph{\textbf{\acrfull{tes}}}
\acrshort{tes} systems such as thermal storage tanks are accumulators that store thermal energy to be used in a later stage when consumption is required or when energy generation is cheaper. They can store energy from the surplus heat generated by industrial processes, the excess renewable production or obtained during periods when electricity production costs are lower by transforming this energy into thermal energy.\\
Thermal energy can be stored using a variety of mediums, but liquid water is the most common in both domestic and industrial applications. This preference is due to its widespread availability, non-toxic nature, high thermal capacity, and the broad range of temperatures within which it can operate \cite{han2009thermal}, we therefore only consider water storage. Using water can also give a secondary role to the system, which can also act as a reservoir. Thermal energy storage systems can store both cool and hot water, which can be contained in either fully stratified or fully mixed water tanks \cite{hasnain1998review, hasnain1998review2}.\\
Thermal storage optimization in buildings has received comparatively less attention in the literature, though some notable studies have addressed this area. For instance, authors of \cite{kircher2015model} apply Model Predictive Control to optimize thermal storage in a commercial building under dynamic pricing, demand charges, and response programs. Using convex optimization, MPC efficiently balances load shifting and cost reduction, outperforming heuristic methods in simulations.

In \cite{FERNANDEZSEARA2007129} and \cite{fernandez2007experimental} authors conducted empirical analysis of static (resp. dynamic) domestic hot water tanks evaluating the performance according to several measures. In their work the total thermal energy $Q$ of the device, which corresponds to the total thermal energy stored is computed from the energy stored in each water layer $j \in J$ of the tank given by equation \ref{eq:dhw-total-energy-store}:
\begin{equation}
    \label{eq:dhw-total-energy-store}
        Q_{st}(t) = \sum^J_{j=1} \left[ Q_j(t)\right]
\end{equation}

where the energy stored in any water layer relative to the water temperature at the beginning of the heating process is given by equation \ref{eq:dhw-energy-store} with $C_p$ specific heat at constant pressure ($J/kg K$), $T$ temperature ($K$), $\rho$ density ($kg/m^3$):
\begin{equation}
    \label{eq:dhw-energy-store}
        Q_{j}(t) = (V \rho C_p)_j (T - T(t=0))_j
\end{equation}

More specifically the energy efficiency of the device can be described with respect to the heating and cooling periods efficiencies. That corresponds to the ratio of the energy available in the tank at any time to the energy supplied by the heating element until the instant considered, with $Q_{ele}(t) = E \times t$, and the energy accumulated in the tank at the beginning of the process:
\begin{multicols}{2}
\begin{equation}
    \label{eq:dhw-heating-efficiency}
    \eta_h(t) = \frac{Q_{st}(t)}{Q_{ele}(t)}
\end{equation}\break
\begin{equation}
    \label{eq:dhw-cooling-efficiency}
    \eta_{c}(t) = \frac{Q_{st}(t)}{Q_{st}(t=0)}
\end{equation}
\end{multicols}
    
Results of the experiment indicate that the energy efficiency of heating is high, with values exceeding 85\%. The efficiency of useful heating energy is primarily influenced by the initial water temperature. However, the exergy efficiencies of the heating processes are notably low, approximately around 5\%. Both heating and cooling processes are impacted by the pressure within the tank, initial temperatures and preceding process.

\subsection{Levels of EMS}
\begin{figure*}[!ht]
    \centering
    \includegraphics[scale=0.5]{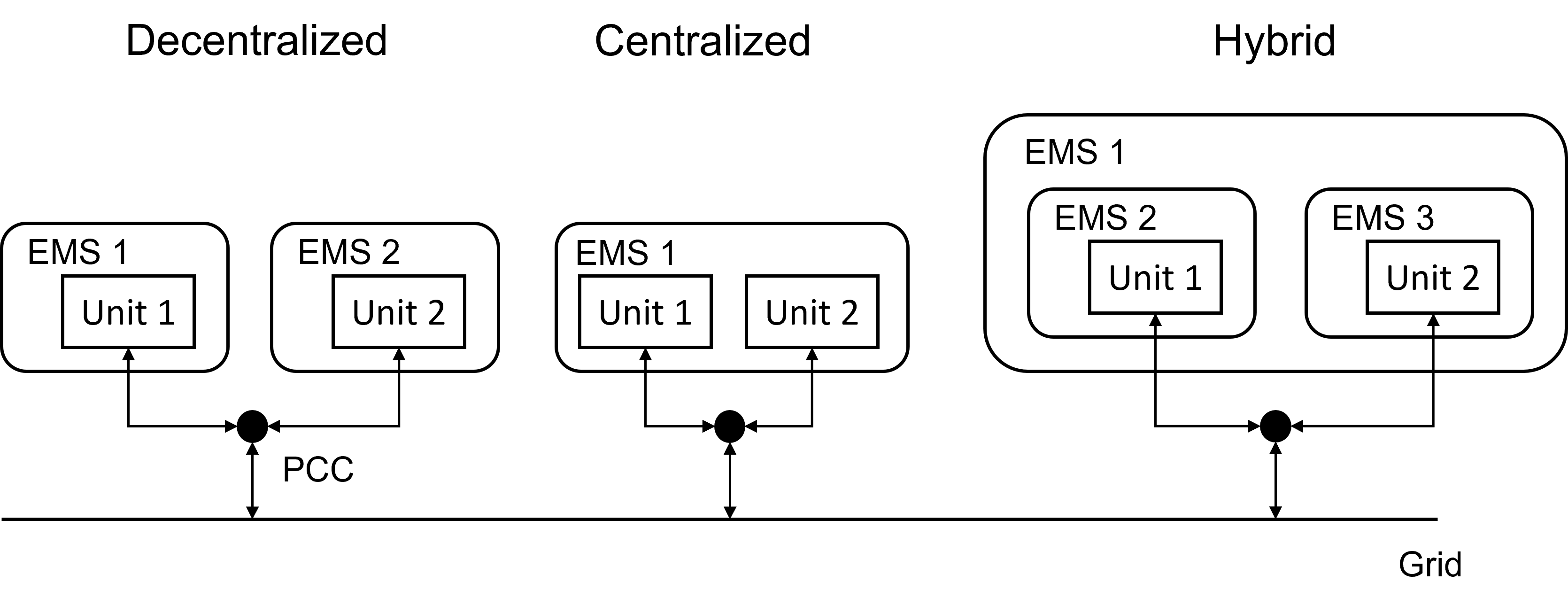}
    \caption{Decentralized, centralized and hybrid EMS representations.}
    \label{fig:ems-level}
\end{figure*}

In the last section we presented the main components in \acrshort{bems} modeling. They can be managed using different strategies that we can broadly classify into 3 categories: centralized which we used in our experiments, decentralized and hybrid EMS described in figure \ref{fig:ems-level} \cite{KHAVARI2020105465, 8308837}.

In the decentralized model, each system is paired with an \acrshort{ems} that solely manages the given system. The units associated with each system are fully autonomous and operate based only on the information of their respective systems, without knowledge of the actions of neighbouring \acrshort{ems}'. These independent systems may need to act collectively to find an optimal solution, as the concatenation of local optimal solutions could lead to a sub-optimal global solution. However, the main advantages of this approach include high reliability and easy scalability. In fact, if one system or unit encounters a malfunction, the rest of the systems would still be able to operate. Moreover, privacy and scaling are not issues in this setting, unlike in the centralized model, as the units do not share information and can easily be added or removed without impacting the overall system.

On the other hand, in the centralized model of management, a single unit collects all data from every system under its control. This data can include a wide variety of factors, including temporal aspects, weather conditions, consumption levels, state of charge, pricing, emissions, and types of load, among others. The centralized unit then makes decisions for each system it manages based on the collected data and predefined objective functions. In this context, the controller can adjust the decisions for each system based on their impact on other systems or the overarching objective, given its comprehensive knowledge of all relevant information. However, this model does not allow for privacy, as each unit must share all of its information with the centralized controller. This lack of privacy could pose a potential drawback for microgrid \acrshort{ems} managing personal buildings or more strategically sensitive structures, such as government buildings. Moreover, as the number of systems managed by the unit increases, the complexity of the model and its computations also escalate, potentially leading to the curse of dimensionality in some cases, which can inhibit the addition of too many systems. This method also presents scalability issues. Each time a new system is integrated, the \acrshort{ems} may need to be recalibrated or retrained, depending on the technical solution employed.

Lastly, hybrid EMS have emerged to combine both of the aforementioned methods. In this kind of setting each unit is dispatched on a system that it manages, but exchanges information such as the action taken to a global controller which concatenates every information and that can act as a regulator.

\section{Related work}
\label{sec:liter}
The field of energy management systems has diverse applications, addressing tasks like controlling voltage frequency, reducing peak consumption, reducing CO2 emissions, reducing energy costs, and ensuring occupant well-being. These approaches can be categorized into different groups among which mathematical and programming control methods, optimisation strategies and RL-based algorithms. 

Control methods involve mathematical programming, classical programming, and selecting variables to optimize a given function within set constraints. These methods utilize both linear and nonlinear optimization models, along with rule-based decision controllers \cite{vuddanti2021review}. The advantage of classical methods lies in their ease of implementation, making them commercially accessible for building applications. They are particularly suitable for integration into existing buildings as some feature low computational demands \cite{taheri2022model}. Optimisation strategies on the other hand are largely based of heuristic techniques. They can integrate multiple objectives while handling non-linearities and discontinuities without risking an explosion of the computational cost. Lastly, RL agents operate in a given environment in order to find an optimal decision policy by trial and error, allowing adaptive learning and optimization, contrasting with the static nature of traditional optimization strategies.

Rule-based control employs expert knowledge through rule-oriented decision-making algorithms. \cite{teleke2010rule} proposes a rule-based control method for managing an energy storage system coupled with renewable sources, addressing intermittence without requiring a mathematical model. In \cite{salpakari2016optimal}, a combination of cost-optimal and rule-based control is presented to minimize electricity costs using market price data and maximizing PV self-consumption over detailed forecasting in real buildings. However, their approach uses simplified models of batteries and \acrshort{tes} and necessitates a precise forecasting module. In \cite{saloux2020optimal}, authors develop a control strategy to control the flow rate of the variable speed pumps in a solar district heating system. Their approach relies on simple heuristic rules and precise tuning of control rule parameters. The main limitation of rule-based methods is their dependency on the defined problem, which means they can not adapt or generalize well on different unseen scenarios. 

Proportional–Integral–Derivative control (PID) control has demonstrated near-optimal performance with sufficient tuning and a well-designed controller \cite{mahmoud2017adaptive}. In \cite{jahedi2011genetic}, authors combine genetic algorithms and fuzzy logic in PID control systems to improve the energy efficiency of dynamic systems. This mixed method benefits from both global and local optimization techniques over specific periods of time. However, it also leads to increased computational demands and longer times for the genetic algorithm to examine all possible fuzzy system characteristics. Additionally, in \cite{yousef2020optimization} the Adaptive Sine Cosine Algorithm (ASCA) is used to adjust PID controller settings tackling the task of controller tuning. However, the addition of Levy flights and adaptive features, to improve the algorithm's search efficiency, also introduces concerns regarding its effectiveness for larger-scale problems.\\
Many studies focus on methods to find optimal parameters as PID control suffers from the sensibility to parameter change and the difficulty of finding optimal combinations of parameters \cite{borase2021review}.

A widely utilized method in energy management optimization is Model Predictive Control (\acrshort{mpc}). \acrshort{mpc} integrates numerical optimization and feedback control, comprising an optimizer solver, a plant model, and a prediction horizon to determine a control trajectory for a process \cite{taheri2022model}. Two main types of \acrshort{mpc} methods exist: physics-based methods, relying on modeling the physics principles of \acrshort{ems} components for optimal solutions, and data-based methods, utilizing historical data to create a statistical model. Authors of \cite{mady2011stochastic} compare stochastic \acrshort{mpc} with a scheduled version, focusing on two main goals: improving occupant comfort and reducing energy use, through predicting building occupancy. This approach tries to find a balance between comfort and energy efficiency. However, its success heavily depends on accurately forecasting occupancy, as they favor energy savings when few people are present. Their approach is therefore dependent on precise occupancy predictions, as errors in these forecasts can greatly affect the trade-off between energy savings and comfort. In \cite{ju2017two}, a two-layer predictive control strategy for energy management in a hybrid micro-grid, equipped with batteries and supercapacitors, considered the degradation cost of storage systems as a short-term factor. The first layer minimized forecast errors, while the second layer optimized operational costs. In \cite{aguilera2018mpc}, an \acrshort{ems} using \acrshort{mpc} and quadratic programming was developed to manage a grid-connected hybrid power plant with renewable energy production and energy storage systems. The approach effectively controlled the energy management system's load cycle while minimizing its impact on its lifespan. While the MPC techniques have good performance in finding the optimal solutions, they also heavily rely on the model accuracy and careful tuning. Moreover, their requirements for heavy computational complexity limit their deployment to real-world systems.

Linear Quadratic Regulator (LQR) methods minimize a specified cost function to derive an optimal policy through the use of weighting matrices provided by an expert. This iterative process aims to find optimal parameters. In \cite{ko2007power}, authors presented a control scheme using a linear quadratic regulator coupled with the Takagi-Sugeno fuzzy model to enhance power quality through a smooth transition of voltage and frequency. In \cite{markovic2018lqr}, an LQR-based optimization technique was proposed for a virtual machine controller in hybrid power systems, adapting parameters based on system frequency disturbance while managing energy consumption. More recently, \cite{abdullah2022linear} utilized an LQR controller to manage a hybrid energy system, aiming to regulate the DC-bus voltage to a known target while maximizing power from a wind turbine. LQR methods are simple and can handle multi-variable systems, however, they rely on strong linear assumptions regarding the operation condition of the system as well as quadratic dependencies between the system's performance and its cost function.

Linear and non-linear programming methods play a significant role in energy management optimization. \cite{lu2015optimal} applied mixed-integer nonlinear programming for scheduling a combined heat and power plant using first-principle models, incorporating constraints such as minimum/maximum power output and steam flow restrictions. In \cite{lauinger2016linear}, authors develop a decision-support tool for managing building energy conversion units and storage devices with a linear program. Their solution demonstrated effectiveness but only adaptable to specific conditions. These techniques both come with certain limitations. Linear programming is limited to linear problems which can prevent its use for more complex and multi-objective settings and is highly sensitive to changes in the objective function's coefficients and the constraints' coefficients. Non-linear programming on the other hand can handle these kinds of problems, but are much more complex to solve due to difficulties in finding a global optimum. Along with high computational costs, it may not always converge to a solution.

\acrfull{flc} is model-independent, relying on previous experience and prior knowledge. While effective in managing multi-criteria control tasks, it is sensitive to hypotheses and initial parameters, involving high computational costs. In \cite{chen2012design}, an \acrshort{ems} for a DC microgrid employed a fuzzy controller to manage battery charge cycles, achieving power equilibrium while maintaining a target battery life. \acrshort{flc} was also applied in \cite{eltamaly2013maximum} to optimize wind turbine energy production, controlling turbine rotational speed for maximum power extraction. In \cite{khalid2019fuzzy}, a fuzzy logic optimization-based controller reduced costs and energy consumption of smart home appliances based on consumption patterns, using heuristic optimization algorithms for shiftable loads.

\acrfull{pso} is versatile and easy to use, requiring minimal hyperparameters. \cite{amer2013optimization} employed \acrshort{pso} to optimize energy generation in a hybrid system, reducing the levelized cost of energy between production and demand load. In \cite{sharafi2014multi}, a multi-objective \acrshort{pso} framework for designing a hybrid power system outperformed similar methods in computational efficiency, considering three distinct objectives using an $\epsilon$-constraint method. 
In \cite{HOSSAIN2019746}, a modified \acrshort{pso}-based energy management method for grid-connected microgrids is presented. The original cost function is modified to better represent battery charging/discharging operations, integrating a dynamic penalty function related to electricity prices. While it operates well in the defined price conditions, it might not adapt well to varying schemes. Like other methods, PSO suffers from scalability and specific tuning issues. It can also struggle in dynamic and stochastic environments, which are the core of energy management problems. Lastly, PSO lacks theoretical convergence guarantees, which might lead to a local optimum.

Reinforcement Learning and Deep Reinforcement Learning techniques train an agent that operates in a given environment in order to find an optimal decision policy by trial and error. Energy management applications are diverse and cover a wide range of different tasks that we can broadly classify in two task group based on the scale at which the controller operates.

\paragraph{Single Device} The first use case is the management of a single device's energy. It focuses on adjusting the input/output loads and activation periods of a given device in order to optimize its usage. Various devices have been studied from \acrshort{ev}s, solar-based water tanks and electric appliances to \acrshort{hvac} systems. They usually intend to minimize the energy consumption of the device and its associated time-varying financial cost and carbon emissions.\\
Different approaches were proposed using a wide variety of RL algorithms, following the progress in the field, for those objectives.
Among the works which have been interested in this type of approach, we can cite the  Gnu-RL approach proposed in \cite{chen2019gnu}. It combines a differentiable \acrshort{mpc} policy to capture domain knowledge and imitation learning to leverage the experience of existing controllers. At the end of the process, the model improves its known policy using the \acrlong{ppo} algorithm. Another interesting approach using Monte Carlo Actor-Critic method with a Long Short-Term Memory (LSTM) recurrent neural network to represent policy and value is proposed in \cite{wang2017long}. Other work chosse to leveraged different algorithms such as \cite{marantos2018towards} who proposed a Neural Fitted Q-Iteration (NFQ) approach, which also utilized a neural network to approximate the Q function or \cite{zhang2018deep} who applied the Asynchronous Advantage Actor Critic (A3C), a deep RL approach, in a simulation of the Intelligent Workplace building in Pittsburgh, USA.\\
Another main focus of single device optimization also considers maximizing their operating cost. Variable-speed wind turbine use control methods to modify the values of parameters that impact their power generation like the tip speed ratio. \cite{wei2015reinforcement} and \cite{wei2016adaptive} used respectively tabular and ANN versions of the Q-learning algorithm to develop maximum power point tracking control algorithms. The agent learns the optimal rotor speed-electrical output power curve from experience and then adapts to the variations of wind speed to maximize the power generation of the wind turbine.
Similarly this use case was extended to photovoltaic array power generation where the energy production depends on irradiation levels, to adapt the perturbation to the operating voltage of the PV array resulting in better efficiency factor for both simulated and real weather data sets while adapting much faster than existing methods \cite{hsu2015reinforcement}.
The optimization of operating parameters also concern the lifetime of batteries. Several authors proposed DQN or DDPG based RL approach to manage the cycle of charge of batteries while taking into account natural degradation that occurs when using electrochemical storage systems \cite{cao2020deep, gorostiza2020deep}.

\paragraph{Multi-level} The second common use case is the management of a micro-grid with several instances of buildings or islands or appliances. The main objective are the same as the single input case but with a specific interest on the interaction between the different instances.
\cite{qazi2020coordinated} studied the concept of energy and reserve scheduling of isolated micro-grid clusters. They use a DQN-based approach to improve the economic performance of micro-grids via jointed energy and reserve sharing among them. In \cite{nakabi2021deep} authors compared empirically seven DRL algorithms to optimize the energy consumption of a micro-grid equipped with different energy sources and with specific priority resources. Their study aimed at demonstrating the effectiveness of RL algorithms in a micro-grid setting.

While those approach demonstrated the benefits of reinforcement learning for energy management systems, they did not investigate the generalization issues of RL-based approach. While looking at the problem from a time-series point of view, we aim a proposing a method to generalize the learning to groups of buildings that share common consumption features.

\section{Problem Formulation}
\label{sec:problem-formulation}
In our study, we examine a set of $N$ buildings denoted as $B = \{b_1, b_2, \ldots, b_{N}\}$. Each building is equipped with a controller responsible for regulating the flow of energy into and out of each \acrfull{esu}. This control mechanism dynamically influences the energy drawn from the grid, increasing the intake when storing energy in the units and decreasing it when releasing previously stored energy.
Within the context of a specific building, let's denote the following variables:
\begin{itemize}
    \item $L_t$: Non-shiftable loads (NSL) in $kWh$ for the given building at time $t$.
    \item $E^{th}_t$: Electrical consumption in $kWh$ associated with thermal needs (e.g., deep hot water, cooling, or heating) at time $t$.
    \item $E^{ESU}_t$: Energy transfers in $kWh$ occurring in an \acrshort{esu}s at time $t$. If a storage unit is storing energy, then $E^{ESU}_t > 0$, and if it is releasing energy, then $E^{ESU}_t < 0$.
    \item $E^{pv}_t$: Energy produced in $kWh$ by renewable sources at time $t$, in our case through solar panels.
\end{itemize}
Additionally, we denote $H_t$ the \acrshort{soc} of any \acrshort{esu}, as the normalized proportion of energy stored in the storage device relative to the device's capacity $z$. The building's total energy consumption is defined by:
\begin{equation}
    \label{eq:net-building-conso}
        E_t = L_t + E_t^{th}+ E_t^{ESU} + E_t^{pv}
\end{equation}

At all times, the building's consumption must be satisfied, achieved through a combination of utilizing self-generated renewable energy, releasing stored energy from storage units, and acquiring energy from the grid. Furthermore, the renewable energy is always prioritised, thus:
\begin{equation}
    E_t \geq L_t +E_t^{th} - E_t^{pv}
\end{equation}
Depending on the action chosen by the controller, the remaining energy $E^{r} > 0$ is acquired from the grid.

\subsection{Objective function}
\label{sec:obective-functions}
Our case study proposes an EMS for a single building with multiple objectives. We focus on managing the charge/discharge cycles of the building's \acrshort{esu} in order to limit energy expenditure while at the same time limiting greenhouse gas emissions. To achieve this, we optimize two key objectives, denoted as $C^1$ (\ref{eq:financial-cost}) and $C^2$ (\ref{eq:carbon-cost}) in dollars, representing the financial cost and the financial equivalent of CO$_2$ emissions associated with the energy procured from the network. Both objectives vary over a specified time sequence $\mathcal{T}$ and depend on the real-time financial and environmental costs of a unit of grid energy, denoted as $c^1_t$ and $c^2_t$ at time $t$.

\begin{multicols}{2}
\begin{equation}
    \label{eq:financial-cost}
    C^1(t) = \sum^{\mathcal{T}}_{t=0}E^{r}_{t}\times c^1_{t}
\end{equation}\break
\begin{equation}
   \label{eq:carbon-cost}
   C^2(t) = \sum^{\mathcal{T}}_{t=0}E^{r}_{t}\times c^2_{t}
\end{equation}
\end{multicols}

\subsection{Reinforcement Learning modeling}
By defining the control of power flows hourly, one can consider the \acrshort{ems} as a sequential decision problem. In order to solve this problem, every building $b_i$ and its dynamical environment can be modeled as a Markov Decision Process (MDP): $M = <S,A,P,R,\gamma>$ in which $S, A, P, R$ are set of states, set of actions, probability of choosing an action in a given state and ending in another state, and reward function respectively, $\gamma \in [0,1]$ is the discount factor. In an MDP, the main objective is to find an optimal policy $\pi: S \longrightarrow A$ that maximizes (minimizes) the expected discounted sum of rewards (penalties):
\begin{equation} \label{eq:rl-goal}
    J = \mathbb{E}_\tau \left[ \sum^{\mathcal{T}-1}_{t=0} \gamma^tr_t \right],
\end{equation}
where $\tau$ is a trajectory generated by policy $\pi$ and $\mathcal{T}$ the maximum episode length.

The \textbf{state space} is described as a vector such as each $s_t \in S$ is defined for a given building as:  
\begin{equation}
    \label{eq:state-space}
    s_t = \left\{C^{\text{g}}_t, \{H_t\}, E_t^{\text{pv}}, L_t, \{K_t\}\right\}.
\end{equation}
It includes the cost of a grid energy unit $C_t^{\text{g}}$ being the aggregation of $c^1{t}$ and $c^2_{t}$, a set of \acrshort{esu}' \acrshort{soc}, the renewable energy production, the building's non-shiftable load and a set of temporal features $\{K_t\}$ representing the month, day and hour.

To ensure the agent’s actions remain safe and physically feasible, we incorporate domain knowledge into the reinforcement learning algorithm by enforcing dynamic constraints on the action space. The agent’s raw control input is a single continuous action $a_t \in [-1,1]$, which translates to a proportion of the \acrshort{esu} capacity that can be charged or discharged at each timestep. These theoretical bounds define the broadest possible action space—merely limiting actions to values that respect the\acrshort{esu}’s capacity. However, they do not account for additional operational and physical constraints that can come from the building’s energy profile, device characteristics, or safety requirements. To address this, we refine the action space at each timestep using a masking approach driven by domain-specific information and system constraints. The valid action space is determined by considering the current state of charge $H_t$, the available renewable energy $E^{pv}$, and the building’s instantaneous energy demand $E_t$. For example, if the state of charge is zero ($H_t = 0$), discharging the ESU is not physically possible, so the action space must exclude negative values. In other situations, if there is surplus photovoltaic energy, the feasible actions shift toward charging the ESU, whereas if the battery is partially full, the agent can choose from a narrower range of options that neither overcharge nor excessively discharge the device.

We can infer the actual valid action space at each timestep based on the nature of the modeling of the different components, as described in Section \ref{sec:bems}, as well as the general physics constraint that exists. It is given at each time step $t$ for an \acrshort{esu} $d$ by:
\begin{equation}
a_{t} = \begin{cases}
 [ 0, \frac{v_z-H_t}{z_d} ] &\text{if $E^{pv} \geq E_t$}\\[1ex]
 [ - \frac{\max(E_t,H_t)}{z_d}, \frac{z_d-H_t}{z_d} ] &\text{else}
\end{cases}
\end{equation}

where $z_d$ is the nominal capacity of the ESU and $v_z$ represents the maximum feasible energy input under the current conditions. By applying this masked, domain-informed action space, the agent is guided to select only those actions that are possible, maintaining equipment integrity, and ensure system safety and facilitating its deployment into real world sytems.

The \textbf{reward function} provides an interface between the agent's policy and goal. The main objective is to learn an optimal policy that decides the optimal storage device usage in each state to minimize the total cost. Our reward shaping strategy splits into two components: one emphasizes the maximization of self-generated renewable energy utilization, while the other concentrates on managing storage cycles during periods when renewables are not at their peak.

\paragraph{Photovolatic based reward}
We designed the reward function as an adversarial reward signal for situations where the renewable production exceeds the total consumption of the building to incentivize the agent to charge the storage unit when free energy is available. As a baseline situation we choose the building's consumption cost without storage nor \acrshort{pv} capacities. The final reward is the difference between the baseline cost $C^b$ and the cost derived from the action of our agent equipped with PV modules and a storage capacity $C^a$.

\begin{equation}
    \label{eq:reward-pv}
    \begin{aligned}
        R(t) & = C^b_t - C^a_t
             & = C^g_t \left( E^{r,b}_t - E^{r,a}_t \right)
    \end{aligned}
\end{equation}
with $E^{r,b}_t$ and $E^{r,a}_t$ the remaining energy coming from the grid in respectively the baseline case and the agent case. In the baseline case, all of the energy is coming from the grid. 

\paragraph{Energy storage system management}
The second part of the reward focuses on managing cycles to effectively reduce operating costs.
The reward function depicts the cost of each unit of energy consumed by the building and is defined based on two costs: 1) $R^1$, associated with energy usage from storage units, and 2) $R^2$, derived from grid energy usage for residual consumption.

The $R^1$ computation requires the cost of the energy unit $C^d$ for the storage device. This cost is maintained by tracking the quantity and associated cost of energy transferred to the device $E^{ESU}$.
\begin{equation}
    C^d_t = \frac{C_{t-1}^d \times H_{t-1} + E^{ESU} \times C^{\text{g}}_t}{ H_{t}}
\end{equation}

Only addressing cost $R^1$ at discharge fails to convey optimal charging time, which might wrongfully induce the agent to maintain a complete charge rather than optimizing charging phases. To address this issue, we introduce hyper-parameter $\zeta$ to the final cost function. It adjusts the cost of using stored energy between charging and discharging periods. A $\zeta$ value of 0 or 1 signifies only discharge or charge time cost consideration, respectively.
\begin{equation} \label{eq:test}
\begin{aligned}
     R(t) & =  R^{1}_t + R^2_t   \\
        & = (1-\zeta) C^{d}_t\left(\max(0, H_{t-1}-H_t)(1-\alpha)\right)\\
        & + \zeta C^g_t \sum_{e \in E^{th}} E_t^r + E^{{ESU}}
\end{aligned}
\end{equation}
The second part of the equation relies on the cost of responding to remaining demands with by using the grid.

\section{Proposed Algorithm}
\label{sec:approach}
\begin{figure*}
    \centering
    \includegraphics[scale=0.6]{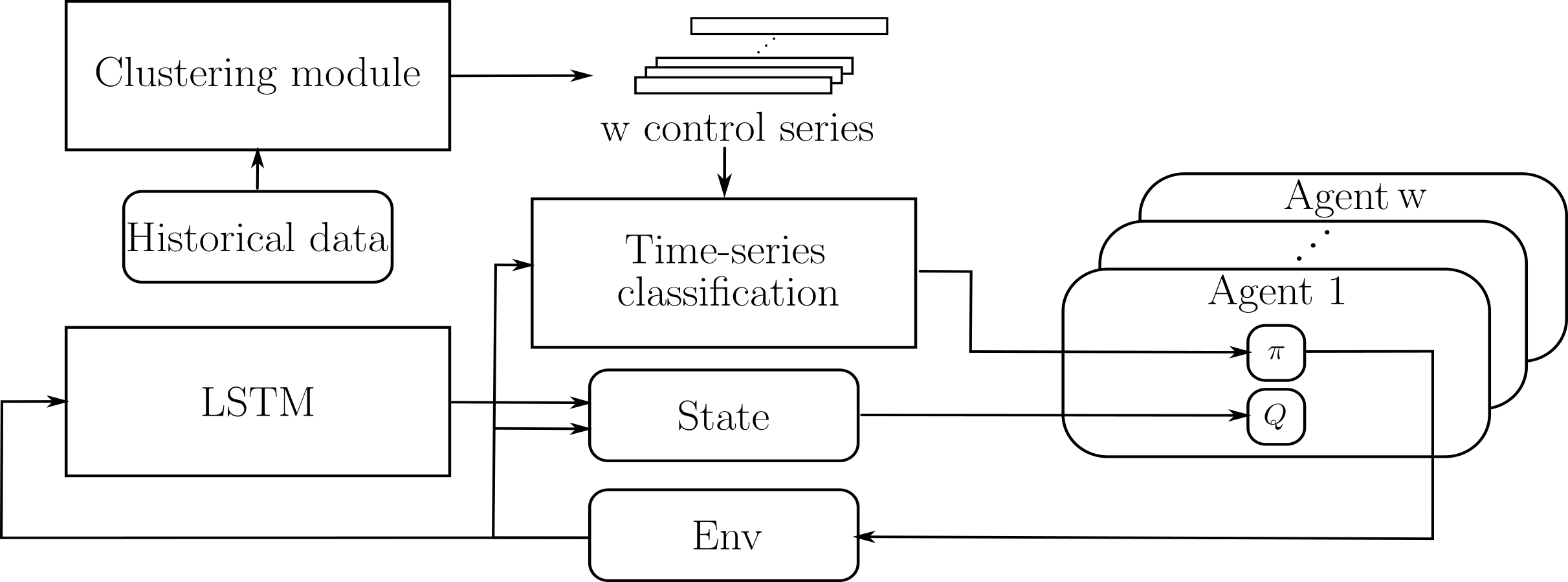}
    \caption{The overall framework of the proposed approach. The clustering module is used on historical data to extract common behaviors. They are utilized during training to map a specific policy to each behavior. The LSTM module enhances the prevision capacities of the agent.}
    \label{fig:framework}
\end{figure*}

An analysis of the training dataset reveals diverse consumption profiles among the buildings, indicating the presence of different building types such as residential units and commercial stores. This range of building types poses a significant challenge: developing a solution that is adaptable across various building groups, each with unique energy consumption patterns and requirements. To address this challenge, we propose a method that can capture and generalize the underlying behaviors of different buildings, enabling effective energy management across this heterogeneous set.

To systematically identify and categorize similar behavioral patterns among buildings, we employ time series analysis coupled with clustering techniques on their consumption loads. Specifically, we focus on the non-shiftable loads, which are portions of energy consumption that cannot be rescheduled or controlled in response to external signals—such as essential heating, ventilation, air conditioning (HVAC), and lighting systems. These loads are representative of a building's intrinsic energy demand and operational characteristics, making them a reliable indicator of overall consumption behavior. The reason for using non-shiftable loads lies in their consistency and availability in historical data. Unlike shiftable loads, which can vary significantly based on user behavior or control strategies, non-shiftable loads provide a stable basis for analysis. By examining these loads, we can capture the energy consumption patterns of different building types.

We preprocess the historical non-shiftable load data to normalize and standardize consumption profiles, accounting for seasonal variations and occupancy patterns. We then use hierarchical agglomerative clustering to group buildings based on similar load profiles, minimizing variance within clusters. By calculating the Euclidean distance between load profiles, we quantify dissimilarities, allowing us to identify distinct clusters that represent different energy consumption patterns.

In addition to clustering, we incorporate both learned and prior domain knowledge into the reinforcement learning algorithm to enhance its performance and ensure safe operation. Firstly, we improve the RL agent's performance by forecasting incoming observations using predictive models. This allows the agent to anticipate future states of the environment, enabling more informed and proactive decision-making that optimizes long-term energy management objectives. Secondly, we restrain the action space of the RL agent based on prior domain knowledge and safety constraints. By limiting the agent's actions to those that are safe and feasible, we prevent potentially harmful decisions that could degrade the lifespan of energy storage components or compromise the safety and stability of the grid.

This integrated approach is crucial for the downstream reinforcement learning algorithm for several reasons:
\begin{itemize}
    \item Policy Generalization: By grouping buildings with similar consumption patterns, we can develop energy management policies that are generalizable within each cluster. This reduces the need to design customized policies for every individual building, enhancing the scalability of our approach.
    \item Transfer Learning: The clusters facilitate the application of trained reinforcement learning agents to new buildings within the same cluster. Since buildings in the same cluster exhibit similar behaviors, a policy optimized for one building can be effectively transferred to others, minimizing retraining efforts.
    \item Targeted Optimization: Understanding the specific characteristics of each cluster allows us to develop optimization strategies that address the unique challenges and opportunities within each group, leading to more efficient energy management.
\end{itemize}

By integrating clustering into our methodology, we bridge the gap between the need for individualized energy management solutions and the practicality of deploying scalable systems.  We create the base for applying generalized reinforcement learning agents to different building groups, facilitating effective and scalable energy management solutions across diverse building types. This methodology not only addresses the heterogeneity in building consumption profiles but also enhances the adaptability and efficiency of the proposed energy management system.

\subsection{Clustering of building consumption profiles using time series analysis}
\label{sec:clustering}
To group consumption into clusters, we leverage a clustering method defined in \cite{LUCZAK2016116}. Let $\mathbf{X} \in \mathbb{R}^{n \times m}$ be the the matrix of time series, where $n$ is the number of series and $m$ is the length of each series. For each time series $\mathbf{x}_i$ in $\mathbf{X}$, we compute the derivative $\mathbf{x'}_i$ as follows, which reflects the trend of the series on all points:
\begin{equation} \label{eq:derivative}
        \mathbf{x'}_i(t) = \frac{d}{dt} \mathbf{x}_i(t)
\end{equation}
We then perform the Fourier transformation \cite{sneddon1995fourier} using an FFT (Fast Fourier Transformation) on each derivative series $\mathbf{x'}_i$ to obtain the frequency domain representation, which we denote $\hat{\mathbf{x}_i}$ given by:
\begin{equation}\label{eq:fourier2}
        \mathbf{\hat{x}}_i(\omega) = \mathcal{F}[\mathbf{x'}_i(t)] = \int_{-\infty}^{\infty} \mathbf{x'}_i(t) \, e^{-j \omega t} \, dt
\end{equation}
We employ the FFT as a feature extractor to accentuate significant frequency-related patterns in the data. By utilizing FFT, the resulting series become less complex, enhancing the efficiency of the Dynamic Time Warping (DTW) algorithm, particularly advantageous when dealing with long time series. This approach enables a targeted focus on patterns within the time series that exhibit pronounced frequency dependencies. It is particularly pertinent in the context of electric consumption in buildings, known for showcasing distinct time-related patterns.

We transform our matrix one last time using Dynamic Time Warping (DTW) algorithm \cite{muller2007dynamic}. This technique enables the comparison of similarity between time series that vary in length, speed or frequency, while limiting the effects of shifting and distortion. With its one-to-many approach, the algorithm matches the indexes of each series with at least one index of the other series, enabling a more elastic matching that can detect common patterns and trends even if they are out of phase. To achieve this, the algorithm seeks to arrange the point sequences by minimizing their distance in order to align the series. There are three main steps \cite{senin2008dynamic}. The first is to construct the distance matrix representing the set of pair-wise distances of two series $X$ and $Y$ in some feature space $\Phi$.
\begin{equation} \label{eq:dtw-distance}
    C_l \in R^{N \times M} : c_{i,j} = \|x_i-y_i\|, i \in [1 : N], j \in [1 : M]
\end{equation}
The second step consists of going through the resulting matrix in order to select the alignment. The algorithm exploits the matrix to minimize the cost between points in the sequences while satisfying boundary, monotonicity and step size conditions. The associated cost function is denoted:
\begin{equation} \label{eq:dtw-path}
    c_p(X,Y) = \sum^L_l=c(x_{n_l}, y_{m_l})
\end{equation}
The final solution is then generated by finding the path with the lowest cost using the Dynamic Programming algorithm. The resulting DTW distance function is noted:
\begin{equation} \label{eq:dtw-final}
    DTW(X,Y) = c_{p*}(X,Y) = \min \left\{ c_p(X,Y), p \in P^{N\times M} \right\}
\end{equation}
The algorithm logic is depicted in Algorithm \ref{alg:time-series}.
\begin{algorithm}[H] 
    \caption{Transform and Compare Time Series}
    \begin{algorithmic}[1] \label{alg:time-series}
        \REQUIRE $\mathbf{M} \in \mathbb{R}^{n \times m}$, a matrix of $n$ time series each of length $m$
        \ENSURE $\mathbf{D} \in \mathbb{R}^{n \times n}$, a matrix of distances between transformed time series
        \FOR{$i=1$ to $n$}
            \STATE Compute the derivative of the $i$-th time series $\mathbf{x}_i$: 
            \STATE $\mathbf{x'}_i[j] = \mathbf{x}_i[j+1] - \mathbf{x}_i[j], \forall j \in [1, m-1]$
            \STATE Compute the Fourier transform of the derivative time series $\mathbf{x'}_i$: 
            \STATE $\hat{\mathbf{x}}_i = \mathcal{F}(\mathbf{x'}_i)$
        \ENDFOR
        \FOR{$i=1$ to $n$}
            \FOR{$j=i+1$ to $n$}
                \STATE Compute the Dynamic Time Warping (DTW) distance between $\hat{\mathbf{x}}_i$ and $\hat{\mathbf{x}}_j$: 
                \STATE $\mathbf{D}_{ij} = \text{DTW}(\hat{\mathbf{x}}_i, \hat{\mathbf{x}}_j)$
                \STATE $\mathbf{D}_{ji} = \mathbf{D}_{ij}$
            \ENDFOR
        \ENDFOR
    \end{algorithmic}
\end{algorithm}

Clusters are built from the resulting matrix using hierarchical clustering. Hierarchical clustering is used to build nested clusters by recursively merging or splitting each of them. Here, we use the so-called agglomerative form, which consists of considering each observation as a single cluster at the start of the process and then merging pairs of clusters at each generation. To build our clusters hierarchically, we need to provide a distance measure between each observation, as well as a linkage criterion that indicates the dissimilarity of the clusters in terms of the point-to-point distances of the observations in the clusters under study. The distance measure chosen is the Euclidean distance calculated on the basis of our matrix defined between two points $a$ and $b$ by: $d(a,b) = \sqrt{(a - b)^2}$. For the linkage criterion, we adopt Ward's method, which aims to minimize the total within-cluster variance at each step of the clustering process. This method considers the variance between merged clusters and selects the pair of clusters whose combination results in the smallest increase in total variance. We choose Ward's linkage criterion specifically for its ability to quantify this increase in variance. It is defined as:

\begin{equation} \label{eq:ward}
\begin{aligned}
    \Delta E =\frac{|A| . |B|}{|A| \cup |B|} \|\mu_A - \mu_B\|^2 = & \sum_{x \in A \cup B}|| x- \mu_{A \cup B}\|^2 \\
    & - \sum_{x \in A}|| x- \mu_{A}\|^2 - \sum_{x \in B}|| x- \mu_{B}\|^2
\end{aligned}
\end{equation}
where $A$ and $B$ are the clusters being considered for merging, $|A|$ and $|B|$ their number of observations respectively, $\mu_A$ and $\mu_B$ their centroids and $\mu_{A \cup B}$ the centroid of the merge cluster $A \cup B$ and $\Delta E$ which quantifies the change in total within-cluster variance resulting from the merger of clusters $A$ and $B$.

By calculating the increase in within-cluster variance for potential merges, it ensures that at each step, we merge the pair of clusters that results in the minimal possible increase in total variance. This approach leads to clusters that are as internally homogeneous as possible, which is essential for:
\begin{itemize}
    \item Enhancing cluster quality, as minimizing variance within clusters improves the cohesiveness and distinctness of the clusters
    \item Ensuring robustness so that the clustering process becomes less sensitive to outliers and noise, resulting in more reliable groupings
    \item Improving subsequent policy training that are directly dependent on the data sources identified by the clustering method
\end{itemize}

\subsection{Classification of new buildings for policy mapping}
\label{sec:classification}
Once the clusters have been defined through hierarchical clustering, we obtain $w$ distinct data models, each corresponding to a cluster representing similar building consumption behaviors. For each cluster  $i$, where $i = 1, 2, \dots, w$, we have learned an optimal policy $\pi_i^*$ for managing energy charge and discharge cycles using reinforcement learning.

When faced with a new, unknown building $b_{\text{new}}$, the challenge arises of determining which cluster it most closely resembles so that we can assign the appropriate pre-learned policy  $\pi_j^*$ to it. Rapidly identifying the correct cluster is crucial to minimize the total operating cost for $b_{\text{new}}$ and to ensure efficient energy management without having to wait to collect data from the building to identify its closest cluster.

To classify the new building into one of the known clusters, we begin by preparing a set of reference time series, one for each cluster. For each cluster $i$, we extract a representative control time series $\mathbf{s}_i$, which could be the centroid of the cluster or a prototypical time series that characterizes the typical consumption pattern within that cluster. These reference time series encapsulate the inherent consumption behaviors of the buildings within each cluster.
We then construct a time series matrix  $\mathbf{M}$ of size $(w+1) \times m$, where $m$ is the length of each time series, and $w$ is the number of clusters. This matrix includes the $w$ reference time series $\mathbf{s}_1, \mathbf{s}_2, \dots, \mathbf{s}_w$ and the time series $\mathbf{s}_{\text{new}}$ of the new building:

\begin{equation}
    \mathbf{M} = \begin{bmatrix}
\mathbf{s}_1 \\
\mathbf{s}_2 \\
\vdots \\
\mathbf{s}_w \\
\mathbf{s}_{\text{new}}
\end{bmatrix}
\end{equation}

Next, we compute the dissimilarity between the new building's time series $\mathbf{s}_{\text{new}}$ and each of the $w$ reference time series as in Algorithm \ref{alg:time-series}. The dissimilarity metric $D(\mathbf{s}_{\text{new}}, \mathbf{s}_i)$ quantifies how different the new building's consumption pattern is from each cluster's characteristic pattern. This metric is calculated using the Dynamic Time Warping (DTW) distance:

\begin{equation}
    V_i = \text{DTW}(\mathbf{s}_{\text{new}}, \mathbf{s}_i)
\end{equation}

where $s_{\text{new}}$ and $s_{i}$ are the values of the time series for the new building and cluster $i$, respectively. This calculation results in a dissimilarity vector $\mathbf{V} = [V_1, V_2, \dots, V_w]$, where each element $V_i$ represents the dissimilarity between the new building and cluster $i$.
Alternatively, if the time series don't have different lengths or don't require alignment due to temporal shifts, we may use the Euclidean distance as the dissimilarity metric:

\begin{equation}
    V_i = D(\mathbf{s}_{\text{new}}, \mathbf{s}_i) = \sqrt{ \sum_{t=1}^{m} \left( s_{\text{new}, t} - s_{i, t} \right)^2 }
\end{equation}

After computing the dissimilarities, we identify the index $j$ of the cluster with the minimum dissimilarity to the new building:

\begin{equation}
    j = \arg\min_{i} V_i
\end{equation}

This means that cluster $j$ has the consumption pattern most similar to that of the new building $b_{\text{new}}$. By finding the cluster with the smallest dissimilarity measure, we effectively classify the new building based on its closest match in consumption behavior. Having identified the closest cluster, we assign the corresponding optimal policy $\pi_j^*$ to the new building. This policy encapsulates the learned strategies for energy management specific to buildings within cluster $j$. By mapping the policy of $\pi_j^*$ to the new building, we leverage the insights and optimization strategies derived from buildings with similar consumption characteristics. This methodology allows us to efficiently classify new buildings and apply the most suitable pre-learned policy without the need to retrain the reinforcement learning agent from scratch. It not only saves computational resources but also ensures that the new building benefits from strategies optimized for similar consumption behaviors, leading to immediate and effective energy management.

It's important to note that the choice of dissimilarity measure $D$ is critical to the accuracy of the classification. While the Euclidean distance is suitable when time series are of equal length and aligned in time, Dynamic Time Warping is more appropriate when time series may be out of phase or have temporal distortions. Other distance measures, such as Manhattan distance or correlation-based measures, could also be considered depending on the characteristics of the data and the specific requirements of the classification task.

To classify the new building quickly, especially in real-time applications, we may use a short initial time window of observations from $\mathbf{s}_{\text{new}}$. As more data become available, the classification can be updated or refined to ensure the best possible policy mapping. This incremental approach allows for continuous improvement in the classification accuracy and the effectiveness of the assigned policy.

By integrating this classification methodology into our energy management system, we enhance its ability to adapt to new buildings and maintain optimal performance across a diverse set of consumption profiles. This method facilitates scalability and adaptability in deploying reinforcement learning agents for energy management in buildings with varying consumption patterns, ultimately contributing to more efficient and sustainable energy use in the built environment.

\subsection{Enhancing agent decision-making through LSTM-based environmental forecasting}
\label{sec:prediction}

In the initial setup of our reinforcement learning framework, the agent has access to values related to energy consumption and generation only after a certain delay. Specifically, it can perceive the state of the environment at a given moment only after that moment has passed. This delay is inherent due to data acquisition and processing times. Such a configuration is suboptimal for learning an effective policy, as the agent is constrained to base its decisions solely on past data, typically from the previous hour. This reliance on outdated information limits the agent's ability to respond to rapid changes in the environment, resulting in suboptimal decision-making.

This constraint becomes particularly problematic when the current situation diverges significantly from the recent past. Factors such as volatile energy prices, sudden shifts in user consumption habits, or rapid changes in solar energy production due to weather conditions introduce non-stationarity into the environment. The agent's inability to anticipate these changes hinders its performance in optimizing energy utilization and costs.
To address this challenge, we have integrated a predictive module, following the work of \cite{zangato2024enhancing}, based on Long Short-Term Memory (LSTM) networks into our model. LSTMs are a type of recurrent neural network (RNN) well-suited for capturing temporal dependencies and long-term patterns in sequential data. By leveraging the LSTM's ability to model complex time-series relationships, we enable the agent to forecast future values of critical environmental variables, such as energy demand, renewable generation, and energy prices.

The LSTM predictive model is designed with 2 hidden layer consisting of 50 neurons, followed by a linear output layer. The architecture balances the model's capacity to learn complex temporal patterns with computational efficiency. The model takes as input:
\begin{itemize}
    \item Temporal Features: Time-of-day, day-of-week, and other cyclical features provided by the environment, which help capture periodic patterns in energy consumption and generation.
    \item Historical Observations: The last $o$ observations of the target variable to be predicted. This window size captures recent trends that are informative for short-term forecasting.
\end{itemize}

Mathematically, the input at time $t$ is represented as: $\mathbf{x}_t = [s_{t-o},\cdots,s_{t-1}, \text{Temporal Fetaures}_t]$ where $s_{t-o}$ denotes the observed value $o$ time steps before time $t$.
The LSTM model is trained using the Adam optimizer with a decaying learning rate to ensure stable convergence and to avoid overfitting. The Mean Squared Error (MSE) is chosen as the loss function, quantifying the average squared difference between the predicted and actual values. The loss function $L^F(\omega)$ is defined as:
\begin{equation}
    {L}^{F}(\omega) = \frac{1}{|\mathcal{B}|} \sum_{i \in \mathcal{B}} \frac{1}{N} \sum_{n=1}^{N} (s_{t+n} - \hat{s}_{t+n})^2,
\end{equation}
where \(s_{t+n}\) is the true \(n\)-th state following the current state \(s_t\), \(\hat{s}_{t+n}\) is the predicted \(n\)-th state, and $|\mathcal{B}|$ is the batch size.

The forecasts generated by the LSTM model are provided to the reinforcement learning agent as additional inputs, effectively augmenting the agent's observation space with predictive information. This integration allows the agent to have a "lookahead" capability, enabling it to anticipate future environmental conditions and make more informed decisions. Incorporating the LSTM predictive module brings several advantages:
\begin{itemize}
    \item Improved Decision-Making: With access to forecasts, the agent can optimize its actions not just for the present moment but also considering anticipated future events, leading to better long-term outcomes.
    \item Adaptability to Non-Stationarity: The agent becomes more robust to changes in the environment, as it can adjust its strategy proactively in response to predicted shifts in energy demand or generation.
    \item Cost and Energy Savings: By anticipating periods of high energy prices or low renewable generation, the agent can strategically manage energy storage and consumption to minimize costs and maximize efficiency.
\end{itemize}

\subsection{Constrained Reinforcement Learning for safe energy management}
\label{sec:algo}

This problem is complex because, depending on each state and situation, there is a constraint on the action space. For example if the building's electric demand is already met by solar production, we cannot discharge the battery. In order to learn a decision-making policy that takes optimal actions, the constraints on the action space should be integrated as a priori knowledge into the \acrshort{mdp}.
We use the maskable version \cite{DBLP:journals/corr/abs-2006-14171} of the \gls{ppo} algorithm \cite{DBLP:journals/corr/SchulmanWDRK17} to address this problem.


The \acrfull{ppo} algorithm acts as a policy gradient method, where an estimator of the policy gradient $\nabla_\theta J(\theta)$ is computed and updated using a stochastic gradient ascent algorithm. It is built on the work of \cite{schulman2015trust} who proposed the Trust Region Policy Optimization (TRPO) algorithm which already implemented the idea of a trust region constraint by indicating that the KL-divergence of the old and new policy should not be greater than a given value $\delta$. The \acrshort{ppo} gradient is of the following form:
\begin{equation} \label{eq:gradient-estimator}
    \hat{g} = \mathbb{E}_t \left[ \nabla_\theta \log \pi_\theta(a_t|s_t)\hat{A}_t \right]
\end{equation}
with $\pi_\theta$ a stochastic policy parameterized by $\theta$ and $\hat{A}_t$ an estimator of the advantage function at timestep $t$ defined by:
\begin{equation} \label{eq:advantage-func}
    A(s_t,a_t) = Q(s_t,a_t) - V(s_t)
\end{equation}
We can thus construct an associated objective function whose gradient corresponds to the policy gradient estimator,
\begin{equation} \label{eq:policy-objective}
    L^{PG}(\theta) = \mathbb{E}_t \left[ \log \pi_\theta(a_t|s_t) \hat{A}_t \right]
\end{equation}
The algorithm uses a singular objective function known as the clipped surrogate objective function that constrains the policy change using a clipping function to avoid destructive large weights updates that are too far from the current policy. In that way, it ensure stable updates:
\begin{equation} \label{eq:cliped-surrogate}
    L^{CLIP}(\theta) = \mathbb{\hat{E}}_t \left[ \min(r_t(\theta)\hat{A_t}, \text{clip}(r_t(\theta), 1-\epsilon, 1+\epsilon) \hat{A_t}) \right]
\end{equation}
with $r_t(\theta)$ being the probability ratio between old and new policies defined as:
\begin{equation} \label{eq:probability-ratio}
    r_t(\theta) = \frac{\pi_\theta(a_t|s_t)}{\pi_{\theta_{old}}(a_t|s_t)}
\end{equation}


To ensure the safety of both the electrical grid and the longevity of energy storage components, it is crucial that the agent operates within a valid and safe action space. Actions that may be theoretically permissible within the broader action space could, in practice, be infeasible or potentially harmful. Therefore, we define a valid action space that restricts the agent's actions to safe and feasible operations. When the agent selects an action that is within the theoretical boundaries of the action space but violates the safety constraints of the valid action space, we project the chosen action onto the closest valid action. However, merely correcting invalid actions post-selection does not enable the agent to learn the dynamic boundaries of the valid action space effectively. Moreover, if the reward function does not penalize invalid actions explicitly, the agent remains unaware of its violations, hindering the learning process.

To address this, we explicitly define the limits $\beta^1$ and $\beta^2$ of our valid space using equations \ref{eq:high-bounds-action-space} and \ref{eq:low-bounds-action-space}. These boundaries are computed based on the physical constraints of the system, including $\phi$, the maximum output power of the source charging the specific \gls{esu} as well as $\Gamma$ and $\eta$ respectively the maximum input/output power and the efficiency of the \acrshort{esu} respectively.

\begin{multicols}{2}
\begin{equation}
    \label{eq:high-bounds-action-space}
    \beta^1 =  -\frac{1}{z_d}\max \left(\Gamma_t,\eta_t H_t,E_t\right)
\end{equation}\break
\begin{equation}
    \label{eq:low-bounds-action-space}
    \beta^2 =  \frac{1}{z_d}\min \left(\Gamma_t,\frac{z_d - H_t}{\eta_t},\phi_t - E_t\right)
\end{equation}
\end{multicols}

Here $z_d$ represents the nominal capacity of the ESU, $H_t$ is the current state of charge at time $t$, and $E_t$ is the energy required to reach the desired state. By incorporating these safety constraints into the action space, we not only ensure safe operation but also enhance the learning efficiency of the agent. By limiting the action space to valid, feasible actions, the agent focuses its exploration on actions that are both safe and effective, avoiding wasted effort on infeasible or dangerous actions. We apply these constraints by discretizing the original continuous action space $[-1,1]$, partitioned into two equal sections representing charging and discharging phases. This discretized action space incorporates the calculated limits $\beta^1$ and $\beta^2$, ensuring all possible actions are within safe operational boundaries.

To further ensure safety and learning efficiency, we employ an action mask during training. This action mask filters out invalid actions by assigning them significant negative logit values in the policy network's output, effectively making their sampling probability zero leading to:
\begin{itemize}
    \item Safe Trajectories: The trajectories $T$ generated during training consist only of valid actions, preserving the safety of system components and the electrical grid.
    \item Enhanced Learning: By focusing solely on valid actions, the agent's policy gradient updates are computed using relevant data, improving learning efficiency and convergence.
\end{itemize}

The policy network, modeled as a neural network, uses a softmax function over unnormalized logits to form an action probability distribution. By masking invalid actions, we guide the agent to explore and learn within the valid action space, improving both safety and performance. This approach not only safeguards the physical components but also simplifies the learning process by eliminating the consideration of infeasible actions.

\section{Experiment} \label{section:experiments-results}

\begin{figure}[!ht]
    \centering
    \includegraphics[scale=1.4]{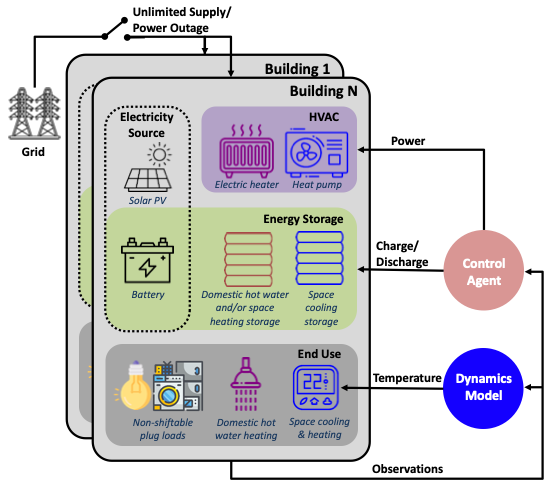}
    \caption{Citylearn environment overview \cite{vazquez2019citylearn}.}
    
    \label{fig:citylearn}
\end{figure}

\subsection{Setting}
Our experimental evaluation is conducted using the CityLearn environment, an open-source, Gym-compatible simulation framework designed for developing intelligent agents that coordinate building-level energy systems and participate in demand response programs within urban settings \cite{vazquez2019citylearn}. Known for its scalability, CityLearn spans over 60 buildings situated in four distinct climatic zones across the USA. The building model in CityLearn adopts a hybrid approach, combining first-principles-based and data-driven models to simulate both thermal and electrical behaviors, accounting for both occupants' preferences and the physical attributes of buildings.

The simulated environment incorporates a range of energy systems, including \acrshort{esu}s and renewable energy devices, each building being uniquely equipped. The central objective within the CityLearn framework is to optimize energy consumption strategies, particularly by controlling the charging and discharging cycles of \acrshort{esu}s in adherence to specific constraints and that pursues cost-effective, low-emission energy utilization.

For our approach, we discretized the original continuous action space $[-1,1]$ into two primary segments—one corresponding to charging and the other to discharging the \acrshort{esu}. Each segment is subdivided into $l=10$ evenly spaced fractions of the \acrshort{esu}’s capacity, representing incremental adjustment steps. In addition, we include an action corresponding to an idle state, thus ensuring comprehensive coverage of feasible operational modes. We partitioned the building dataset into training and testing subsets, ensuring that each set contains a representative mixture of building types and energy profiles for robust evaluation.

To improve learning performances, we normalized the observations fed to the agents. For normalization, we used $8760*5$ observations collected via random agents to center and reduce each new observation during training. Agents took random actions, dictated by a uniform law $U(a,b)$, where $a$ and $b$ are the valid action space bounds.\\
Temporal variables often exhibit cyclical patterns (e.g., daily or seasonal cycles). To preserve and leverage these inherent periodicities, we apply a sine-cosine transformation to each temporal feature. Given a temporal feature $f$ with a known maximum range $f_{\max}$, we replace $f$ with two derived features:
\begin{multicols}{2}
\begin{equation}
    f^1 = \cos(2\pi f / f_{max})
\end{equation}\break
\begin{equation}
    f^2 = \sin(2\pi f / f_{max})
\end{equation}
\end{multicols}

This representation maps the temporal dimension onto a unit circle, ensuring that large numerical gaps at the boundaries (e.g., between hour 23 and hour 0) no longer appear as discontinuities. By maintaining the cyclical continuity of time, the agent can more effectively discern and exploit periodic patterns in energy demand, solar production, and other time-dependent factors.

\subsection{Results} \label{section:results}
\subsubsection{Clustering and classification}

\begin{figure*}
    \centering
    \subfloat[\centering Silhouette score analysis for optimal number of clusters in hierarchical clustering.]{{\includegraphics[scale=0.4]{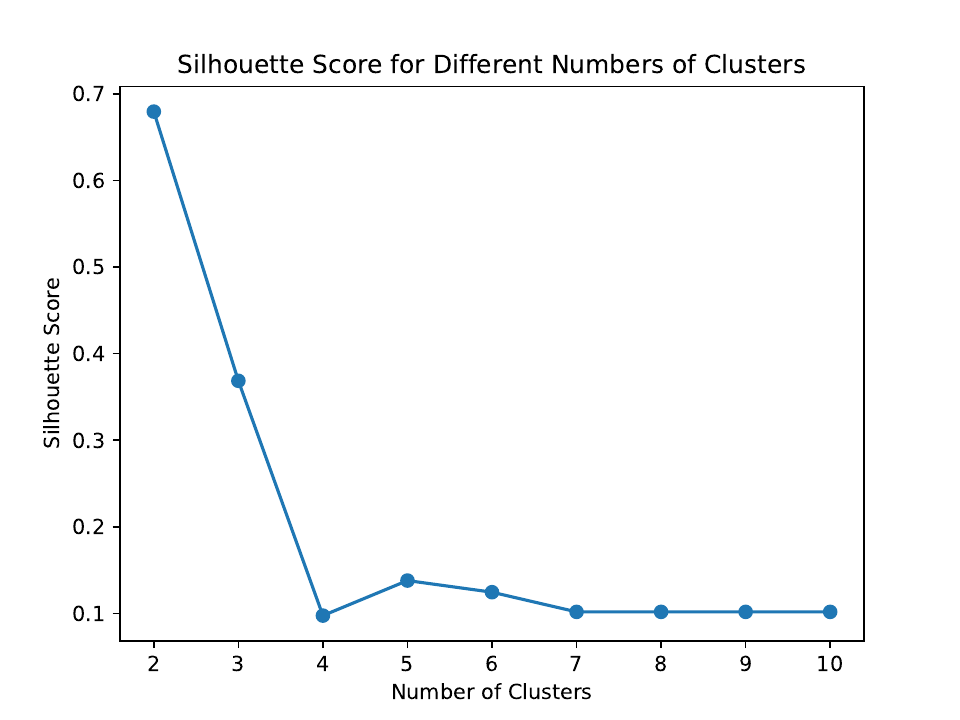}}}%
    \qquad
    \subfloat[\centering Inconsistency analysis for determining optimal number of clusters in hierarchical clustering using depth of 2.]{{\includegraphics[scale=0.4]{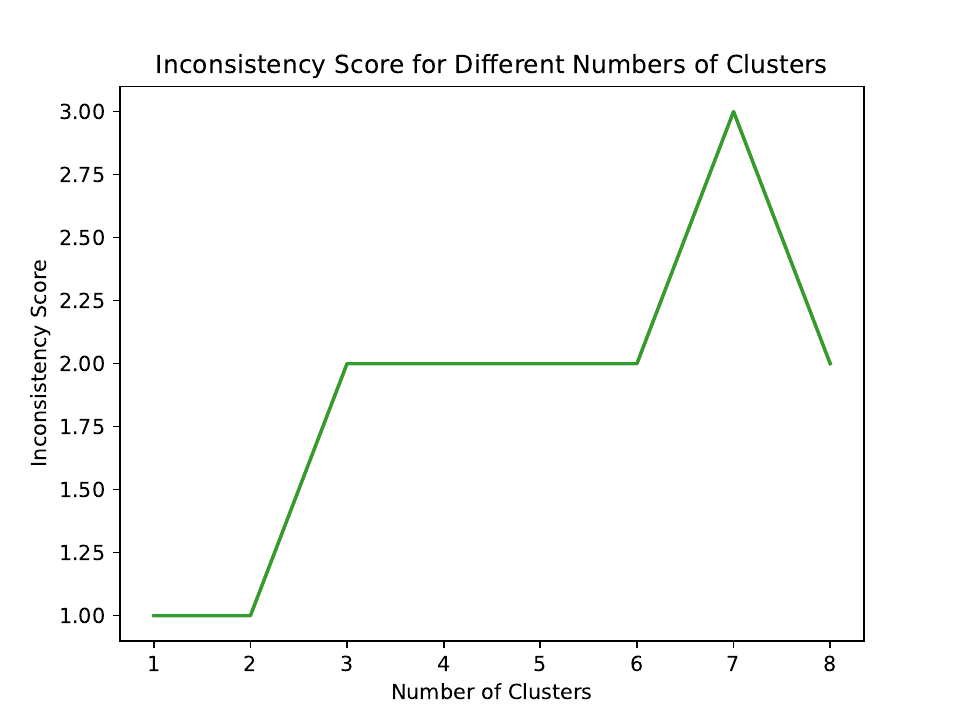} }}%
    \caption{Cluster Evaluation: Silhouette Scores vs. Inconsistency Scores for Different Numbers of Clusters}%
    \label{fig:cluster-analysis}%
\end{figure*}

\begin{figure}
    \centering
    \includegraphics[scale=0.3]{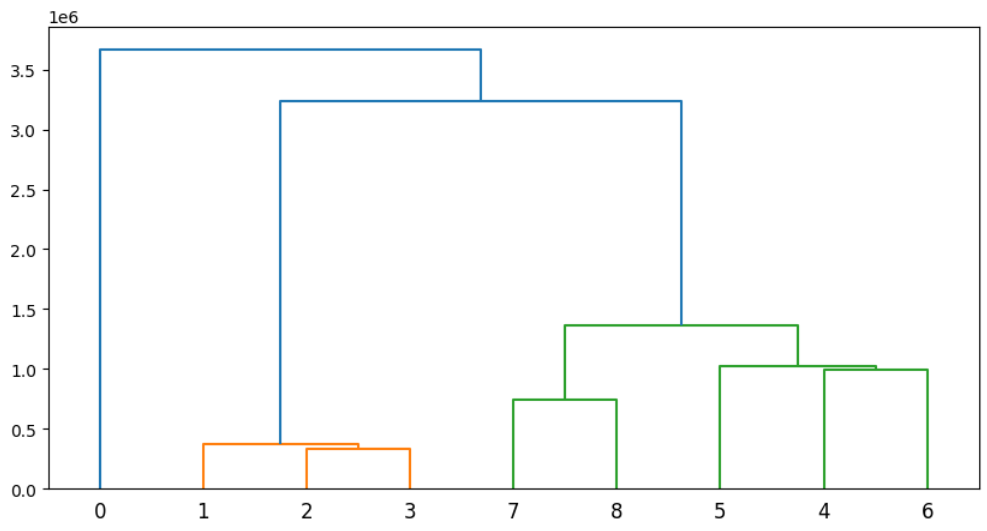}
    \caption{Hierarchical cluster construction dendogram. Each cluster are sequentially merged during the process maximizing the distance between the clusters by maximizing the length of vertical lines connecting nodes.}
    \label{fig:dendogram}
\end{figure}

The first step in our methodology involves identifying latent consumption patterns among buildings by clustering their annual Non-Shiftable Load (NSL) time-series (see Section \ref{sec:clustering}). The NSL serves as a robust and representative metric of baseline energy usage, as it closely correlates with occupancy-driven thermal demands and is consistently present across all buildings. Unlike loads linked to optional devices (e.g., cooling systems), NSL patterns offer a stable and omnipresent reference for identifying consumption behaviors.\\
We determined the number of clusters $w=3$ through a joint analysis of silhouette and inconsistency scores, shown in Figure \ref{fig:cluster-analysis}. Although a pure silhouette-based criterion might suggest two clusters, the inclusion of inconsistency metrics and domain knowledge regarding building diversity influenced our decision. We observed a plateau in the inconsistency measures up to six clusters, indicating diminishing returns in further subdivision. Consequently, three clusters strike a suitable balance, capturing nuanced differences in consumption profiles while maintaining a coherent hierarchical structure also enabling more tailored and specific resulting policies. The dendrogram in Figure \ref{fig:dendogram} further supports the existence of more than two consumption types, aligning with expert intuition and the complexity of real-world building usage patterns.

This clustering solution is not simply an exercise in pattern discovery; it is a crucial step for the downstream reinforcement learning algorithm. First, by grouping buildings with similar consumption profiles, we enable policy generalization. Since each cluster encompasses a coherent set of buildings sharing fundamental consumption traits, a single learned policy can be developed and applied to every building within that cluster. Our results show that clustering effectively segments the dataset into meaningful groups, reducing the need for building-specific policies and enhancing scalability.
Second, our findings highlight the potential for transfer learning. Once a policy is optimized for a representative building in a given cluster, it can be readily transferred to any new building whose consumption profile is identified as belonging to that same cluster. Our empirical results, including the stable clustering patterns across a variety of buildings, underscore this capability. By assigning new buildings to the established clusters, the approach avoids the cost and effort of retraining agents from scratch, thus streamlining policy deployment.
Third, the distinct clusters facilitate targeted optimization. Each cluster represents a unique operational context—variations in user habits, building thermal inertia, or installed energy systems—that can be addressed by a specialized policy. The clustering results confirm the presence of multiple consumption types, allowing the RL policies to focus on each cluster’s specific challenges and opportunities, rather than relying on a one-size-fits-all solution. In this way, the approach improves the agent’s ability to manage energy storage and orchestrate loads efficiently under diverse conditions.

\begin{figure*}
    \centering
    \subfloat[\centering 1 year]{{\includegraphics[width=7.5cm]{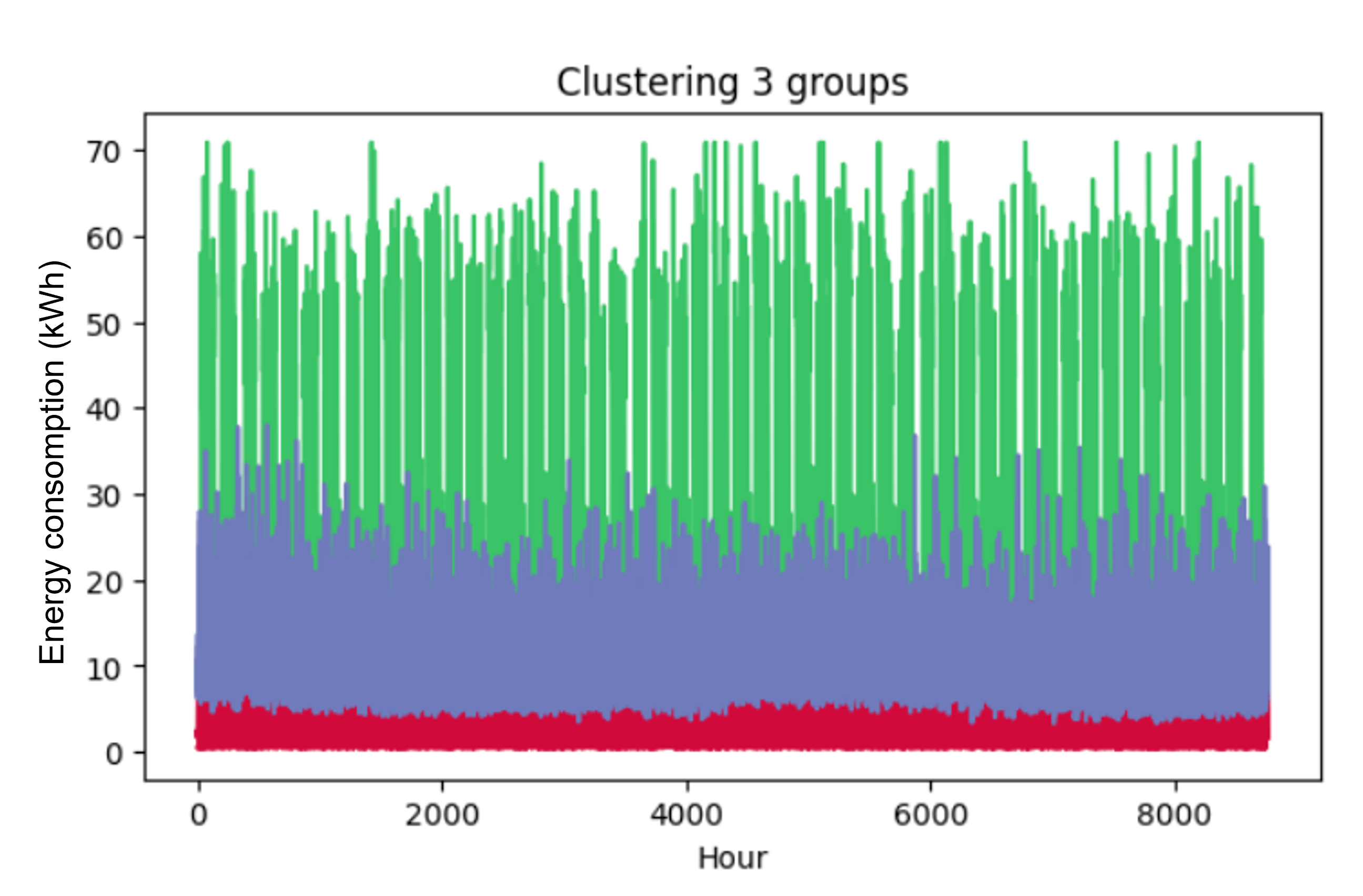} }}%
    \qquad
    \subfloat[\centering 1 week]{{\includegraphics[width=8cm]{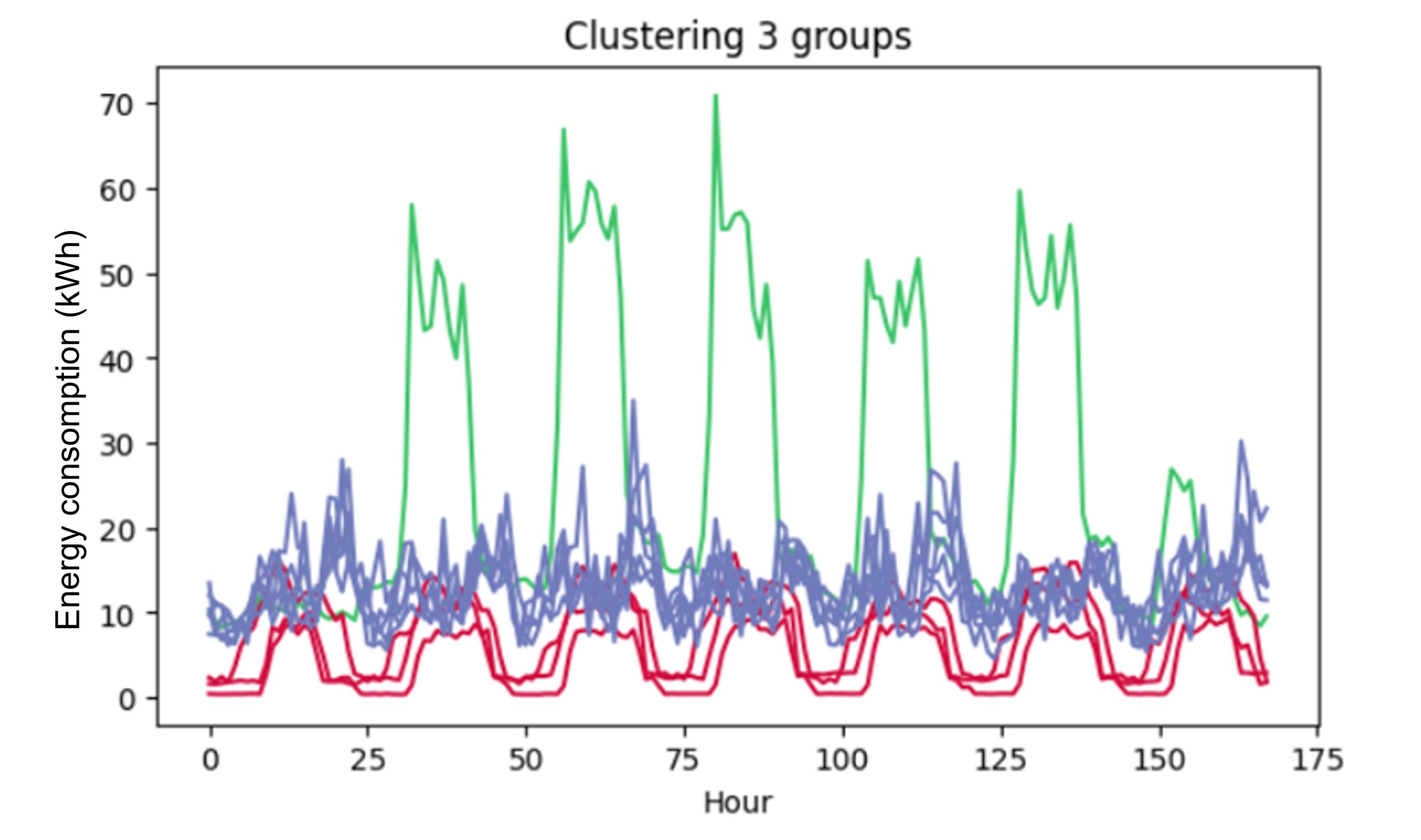} }}
    \caption{w=3 group clustering of a subset of buildings using the clustering method applied on a) 1 year (m=8760) and b) 1 week (m=168) time series. Our 3 policy subtasks are defined based on those 3 different consumption patterns of non shiftable loads in kWh.}%
    \label{fig:clustering}%
\end{figure*}

The robustness of our clustering methodology is further demonstrated by its applicability over different data horizons. To classify a previously unknown building, we reduced the observation period to just $m=7 \times 24$ hours (one week or 1.9\% of a year). Remarkably, the same cluster structure emerged from this brief snapshot, closely mirroring that identified using the entire year’s data. As illustrated in Figure \ref{fig:clustering}, these consistent results validate that even a short observation window can accurately determine the building’s closest consumption type.\\
By rapidly classifying a new building after only one week’s observation, we can immediately assign the corresponding pre-trained policy that best aligns with its observed consumption pattern. In doing so, the agent can start making informed decisions—optimizing energy storage, reducing operational costs, and mitigating environmental impacts—far sooner than if it had to wait for extended data collection or retraining. This capability exemplifies how the clustering and classification module directly supports the overarching objectives: it ensures that policies are generalizable, readily transferable, and specifically tailored to the operational context of each building group.

\subsubsection{Forecasting}
Our LSTM-based prediction module focuses on forecasting key environmental and operational variables—specifically solar energy production, financial costs, and the building’s non-shiftable load (NSL). By predicting these future states, the module equips the reinforcement learning (RL) agent with foresight, allowing it to plan actions that extend beyond immediate conditions and enhance overall energy management. Table \ref{tab:prediction-results} presents a quantitative comparison between our LSTM-driven forecasts and a baseline method that relies on a simple one-hour lag observation. Across the predicted variables, the LSTM-based forecasts achieve lower prediction errors, demonstrating improved accuracy and reliability. This enhanced predictive performance directly underpins the following three objectives:
\begin{itemize}
    \item Improved Decision-Making: The reduction in forecasting error for solar production and NSL signals gives the agent a clearer understanding of the near-future energy landscape. Our forecasts provide the agent with more precise anticipations of when renewable output will decline or when consumption may surge. This predictive capacity enables the agent to make more strategic energy storage and dispatch decisions, ultimately leading to better-informed, long-term policies. The improvement in accuracy translates into the agent’s ability to optimize not just for the present moment, but also for upcoming conditions.
    \item Adaptability to Non-Stationarity: Real-world building energy environments often exhibit volatility, from fluctuating solar irradiance to variable occupant behavior. Our improved forecasts show that even as conditions change over time, the model can capture emerging trends and patterns. For example, when renewable generation patterns shift due to weather changes, the agent no longer relies solely on outdated or lagged observations. Instead, it adapts its strategy proactively, leveraging the LSTM’s predictive insights to maintain stable performance under evolving conditions.
    \item Cost and Energy Savings: The improved accuracy in predicting financial costs and renewable availability allows the agent to strategically time energy purchases, ESU charging, and load shifting. By anticipating periods of elevated prices or reduced solar input, the agent can preemptively store energy when it is cheaper and cleanly generated, subsequently using it during higher-cost intervals.
\end{itemize}

By reducing uncertainty about the future, the LSTM predictive module enables the agent to make better decisions, adapt to changing conditions, and ultimately achieve greater cost and energy savings. These findings validate the importance of incorporating forecasting capabilities within RL-driven building energy management systems and highlight the pivotal role of accurate predictions in achieving sustainable and economically favorable outcomes.

\begin{table}
\caption{Evaluation metrics comparison of LSTM predictor vs. 1-hour lag observations (Base).}
\begin{center}
\begin{tabular}{|c|c|c|}
\hline
\textbf{Observation}&\multicolumn{2}{|c|}{\textbf{Evaluation results}} \\
\cline{2-3} 
\textbf{predicted} & \textbf{\textit{LSTM}}& \textbf{\textit{Base}}\\
\hline
Financial cost & \textbf{0.007}$^{\mathrm{a}}$, \textbf{0.995}$^{\mathrm{b}}$ & 0.702, 0.5 \\
Solar Generation & \textbf{0.128}, \textbf{0.952} & 0.56, 0.798\\
Non Shiftable Load & \textbf{0.117}, \textbf{0.964} & 0.220, 0.871\\
\hline
\multicolumn{2}{l}{$^{\mathrm{a}}$RMSE.$^{\mathrm{b}}$ $R^2$.}
\end{tabular}
\label{tab:prediction-results}
\end{center}
\end{table}

\subsubsection{Policy}

We implemented a reinforcement learning approach to optimize building-level energy decisions, with policies tailored to the distinct consumption patterns identified through clustering. Specifically, we trained three agents—one per cluster—enabling each agent to develop a policy optimized for the characteristic consumption profile of its assigned group. During training, we randomized the building selection within each cluster for every episode, ensuring that the learned policy generalizes within that cluster rather than overfitting to a single building.

\begin{figure}[!]
    \centering
    \includegraphics[scale=0.45]{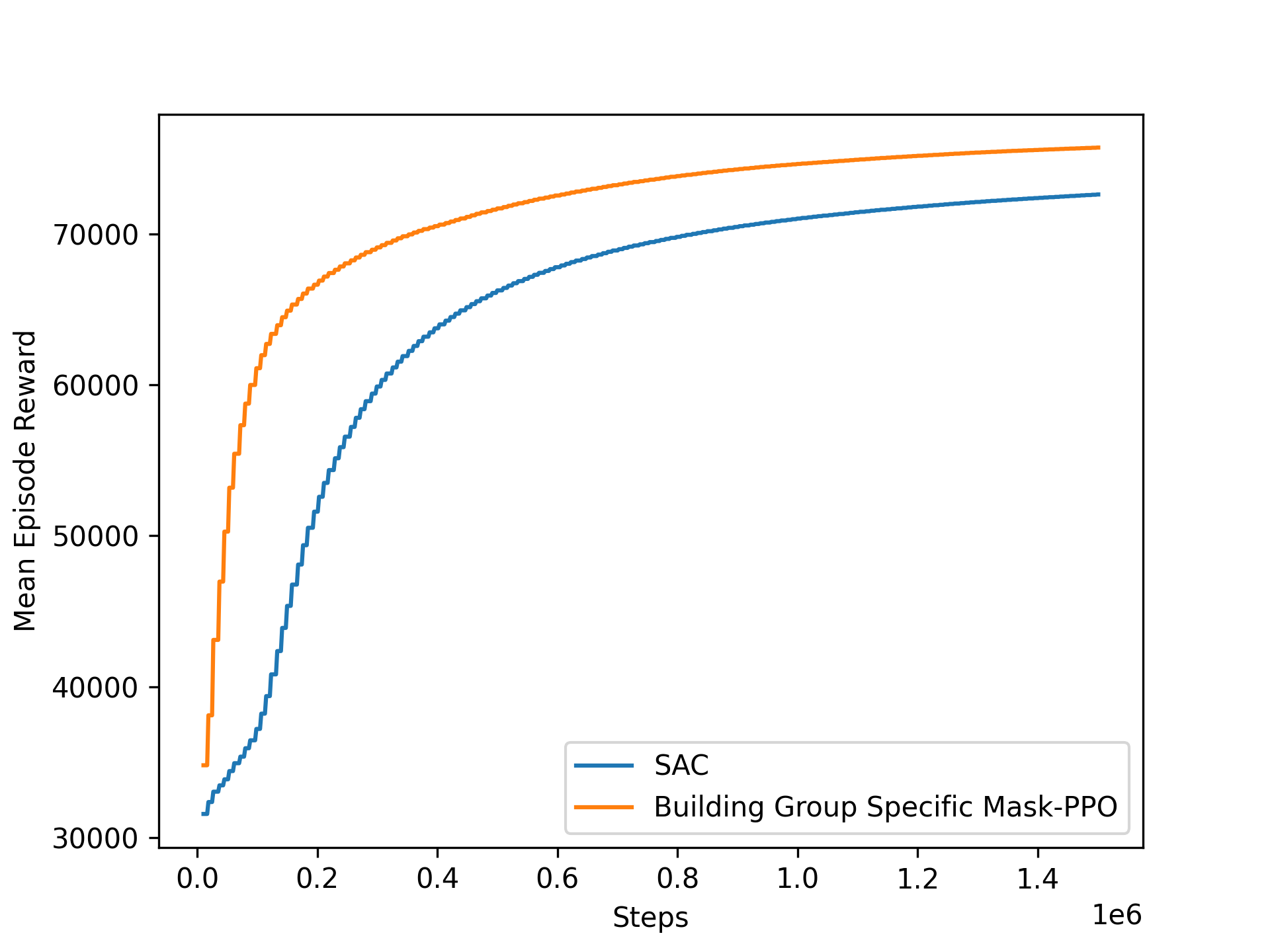}
    \caption{Convergence of the two algorithms compared on a given group of buildings.}
    \label{fig:convergence}
\end{figure}

Our RL-based strategy treats energy management as a sequential decision-making problem, where the agent learns to balance short-term actions against long-term cost and efficiency objectives. On a year-long test set (T=8760 hours), this framework achieved a 5\% reduction in operational costs compared to scenarios with no energy storage. Moreover, benchmarking our results against both a continuous RL algorithm \cite{haarnoja2018soft} and a rule-based heuristic underscores the RL agent’s superiority in navigating the high-dimensional, stochastic energy landscape. Figure \ref{fig:convergence} illustrates the convergence behavior, demonstrating that our approach can discover stable and effective policies more efficiently than other methods. Results in Table \ref{tab1} show metrics for carbon emissions and building operating costs over duration $T$, normalized relative to scenarios without storage. A value of 1.1 indicates a 10\% consumption increase.

Unlike unconstrained methods or naive random strategies, our RL agents operate within a carefully defined and masked action space, guaranteeing that all chosen actions are feasible and align with the operational limits of the \acrshort{esu}. By applying domain-specific constraints, the agent avoids actions that could harm the physical components or destabilize the electrical grid. This constraint enforcement is reflected in the performance metrics: a random policy, which lacks such safeguards, results in hasty and inefficient energy storage management, ultimately increasing costs. In contrast, our agent’s policies systematically leverage the \acrshort{esu}’s capacity within safe boundaries, ensuring reliability and equipment longevity.

The three trained agents, each corresponding to one of the identified consumption clusters, can be seen as “expert” policies tailored to distinct building types. By matching an unknown building’s behavior to one of these clusters, we ensure that the building receives a policy optimized for its particular demand patterns. The result is targeted optimization that caters to each cluster’s unique characteristics, consistently driving costs down by up to 15\% for certain building types while maintaining stable environmental metrics. This stability arises because the agent anticipates energy price variations and schedules \acrshort{esu} operations accordingly, reducing inefficient cycling and thus mitigating unnecessary environmental impact.
these policy results highlight the strength of RL as an optimization method in complex energy settings. By integrating safety constraints directly into the action space and tailoring distinct policies to each cluster’s unique profile, the approach achieves both cost-effective and sustainable energy management. The agent’s well-structured policy execution, stable economic gains, and effective handling of energy price volatility underscore the value of combining RL techniques with domain knowledge and clustering insights to guide decision-making in dynamic building energy ecosystems.

\begin{table*}[!t]
\caption{Algorithm evaluation according to objective functions on test set buildings.}
\begin{center}
\begin{tabular}{|c|c|c|c|c|}
\hline
\textbf{cluster}&\multicolumn{4}{|c|}{\textbf{Evaluation results with compared algorithms}} \\
\cline{2-5} 
\textbf{group} & \textbf{\textit{Ours}}& \textbf{\textit{SAC}}& \textbf{\textit{Random}} & \textbf{\textit{RBC}}\\
\hline
1 & \textbf{1.0}$^{\mathrm{a}}$, \textbf{0.937}$^{\mathrm{b}}$ & 1.018, 0.951 & 1.159, 1.163 & 1.004,0.998\\
2 &\textbf{ 1.013}, \textbf{0.858} & 1.029, 0.876 & 1.186, 1.171 & 1.017, 0.998\\
3 & 1.02, \textbf{0.887} & 1.018, 0.935 & 1.071, 1.062 & \textbf{1.004}, 0.999\\
\hline
\multicolumn{4}{l}{$^{\mathrm{a}}$yearly carbon emissions.$^{\mathrm{b}}$Yearly price cost.}
\end{tabular}
\label{tab1}
\end{center}
\end{table*}

\subsection{Further Analysis: Stochastic Tariffs}

In real-world energy markets, pricing structures and associated environmental costs are subject to continuous change. Grid operators often modify Time-of-Use tariffs to reflect varying supply-demand conditions, encourage load shifting, and manage grid stability. These adjustments can occur frequently and unpredictably, challenging static or rigid control strategies. To evaluate our solution’s adaptability and resilience to such dynamic tariff modifications, we subject the trained RL agent to evaluation scenarios featuring stochastic variations in both financial and environmental cost signals.\\
Our objective is to assess how effectively the agent’s policy, learned under one set of conditions, can cope when the underlying cost structure evolves. Our second objective is to confirm that our approach maintains safe and effective control decisions without requiring retraining from scratch. By adjusting the action masking rules to new constraints if needed, we retain the original policy’s core decision-making logic while seamlessly accommodating tariff changes, thus illustrating the policy’s transferability and scalability.
\begin{figure}[!]
    \centering
    \includegraphics[scale=0.4]{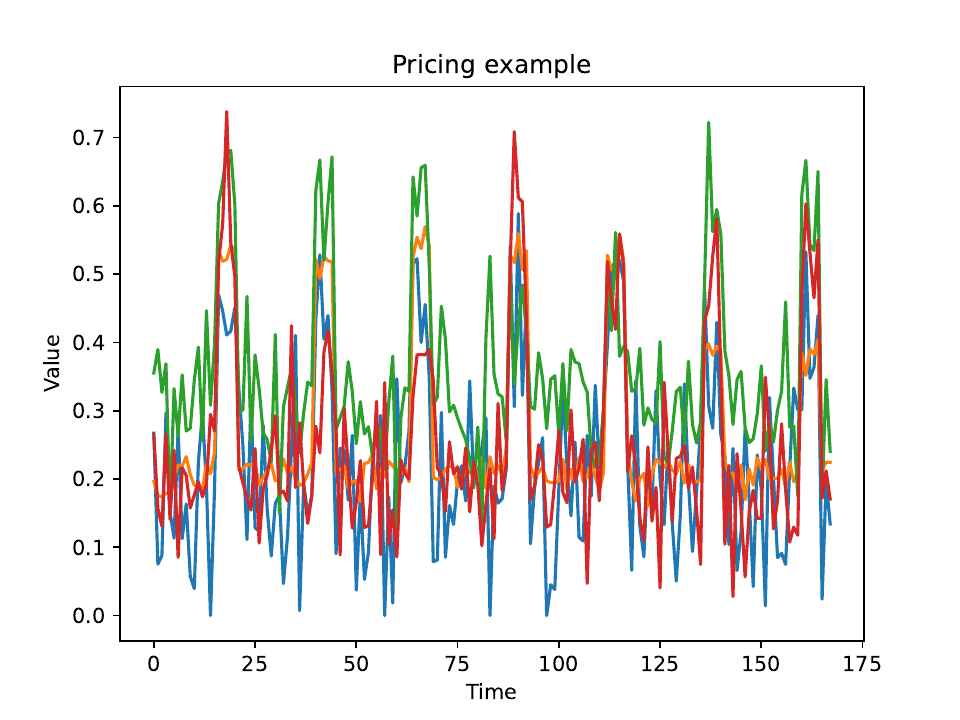}
    \caption{Example of pricing time-series generated for evaluation. Each series represents an example of the pricing of a kWh of energy for a week.}
    \label{fig:pricing}
\end{figure}

To simulate realistic tariff fluctuations, we perturb the original time series of kWh prices and CO$_2$ costs with Gaussian noise. Let $\mathbf{X} = \{x_1, x_2, \ldots, x_m\}$ represent the baseline pricing or cost time series. We introduce noise by sampling $\mu$ and $\sigma$ from uniform distributions $\mathcal{U}(a_{\mu}, b_{\mu})$ and $\mathcal{U}(a_{\sigma}, b_{\sigma})$, respectively. We then generate $\epsilon_i \sim \mathcal{N}(\mu, \sigma^2)$ and form:
\begin{equation}
    \mathbf{X}_{\text{noisy}} = \{x_1 + \epsilon_1, x_2 + \epsilon_2, \ldots, x_m + \epsilon_m \}
\end{equation}

This procedure systematically introduces variability, reflecting uncertainty in grid conditions, policy changes, or unforeseen market events.
In addition, Time-of-Use tariffs often designate high-cost periods to incentivize off-peak load shifting. These periods can be influenced by seasonality, weekdays versus weekends, or even regulatory interventions. To capture this complexity, we randomly select days and time intervals to represent high-cost periods, preserving the underlying logic but injecting stochasticity. As a result, our evaluation environment no longer mirrors the exact conditions under which the agent was trained, but rather a more dynamic and realistic scenario (see Figure \ref{fig:pricing} for an example of modified pricing time series).

\begin{figure*}%
    \centering
    \subfloat[\centering SOC (blue) vs. net electricity consumption (orange)]{{\includegraphics[scale=0.4]{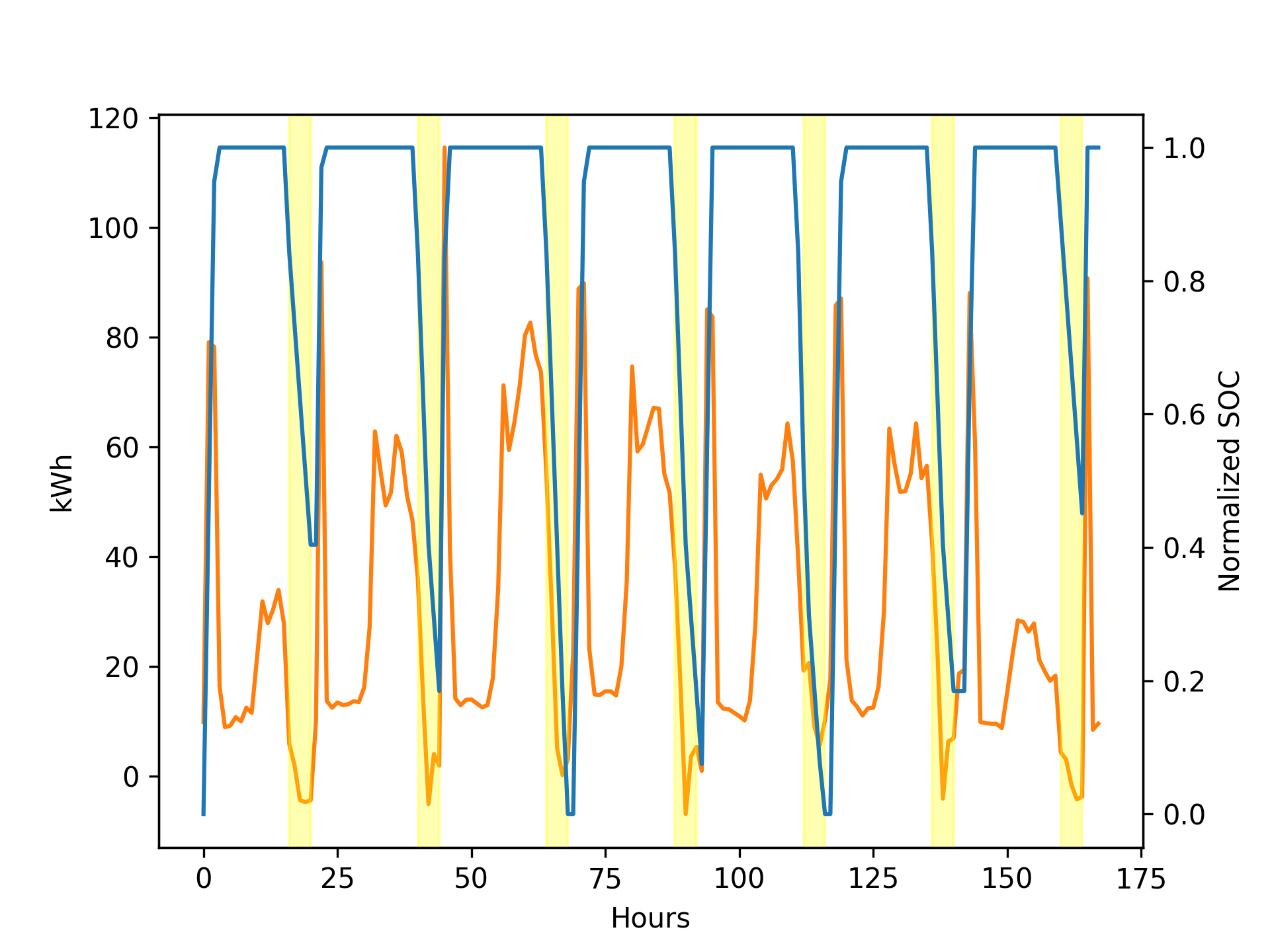} }}%
    \qquad
    \subfloat[\centering Original (orange) vs. net electricity consumption (blue)]{{\includegraphics[scale=0.4]{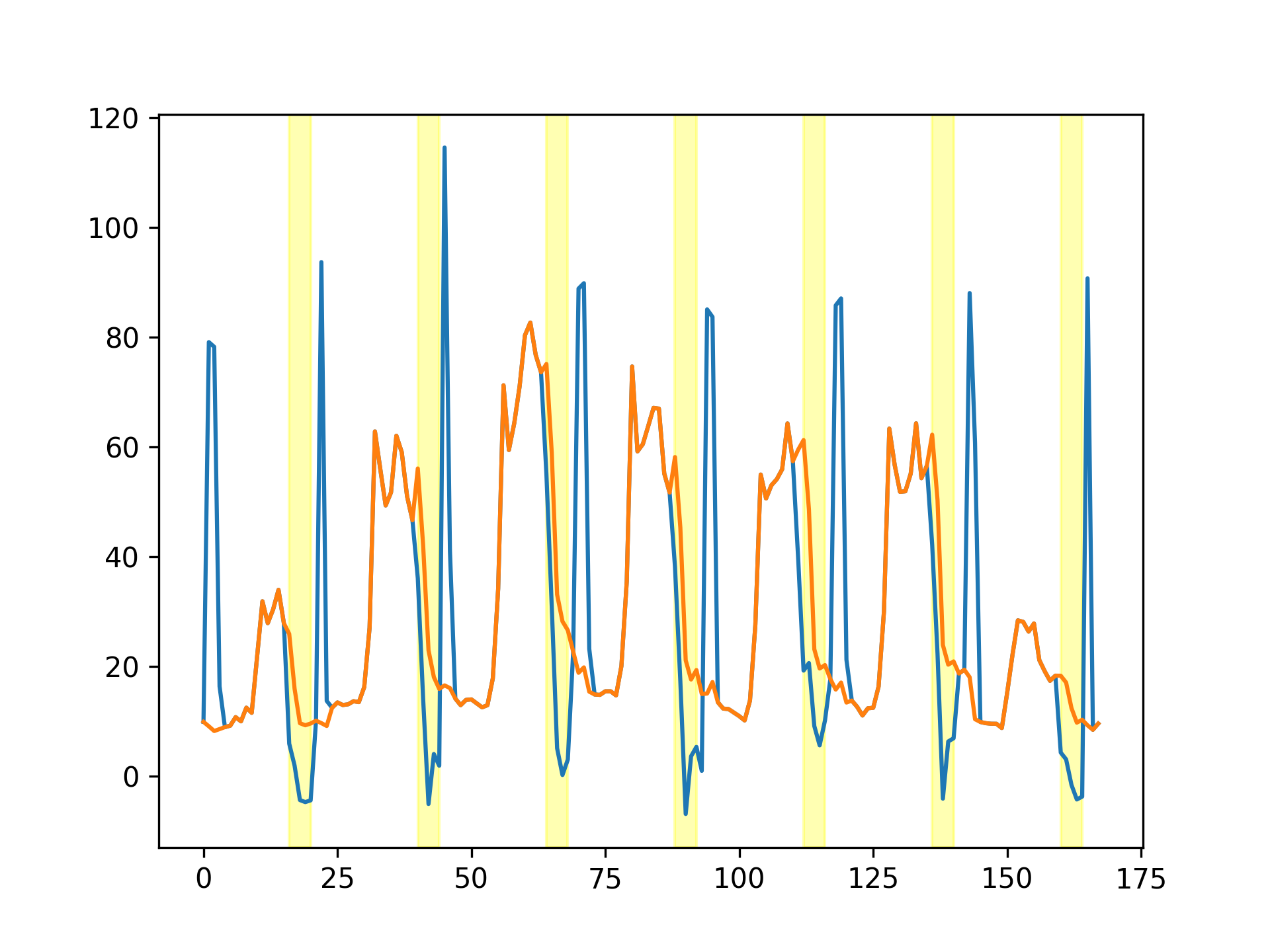} }}%
    \caption{Agent's behavior on an unseen building. Yellow spans represent high Time-of-Use tariffs.}%
    \label{fig:test-agent}%
\end{figure*}

Our empirical evaluation, summarized in Table \ref{tab:results-pricing}, demonstrates that the agent, despite being trained under a particular tariff regime, retains robust performance when exposed to these newly perturbed cost structures. By leveraging the previously learned policies and the ability to refine action constraints post-training, the agent continues to make cost-effective and environmentally stable decisions without retraining from scratch. Figure \ref{fig:test-agent} represents the outcome of the policy by the trained agent of a given building of group 1.
The results highlight several key points:
\begin{itemize}
    \item Adaptability to Changing Conditions: Even when tariffs deviate from their training distribution, the agent effectively adjusts its ESU charge/discharge patterns. This adaptability is a direct consequence of the RL framework’s generalization capabilities and the informed masking of infeasible actions.
    \item Safe Reinforcement Learning Under Uncertainty: The action masking, informed by domain knowledge and system constraints, remains valid even under the new conditions. Thus, the agent avoids harmful actions and respects the updated operational limits.
    \item Transferability and Reduced Overhead: We did not need to re-optimize or retrain the entire policy to handle these tariff modifications. By simply adjusting the stochastic tariffs and potentially refining the action-space constraints, the existing policy readily adapts. This capacity for on-the-fly adaptation underscores the approach’s scalability and cost-effectiveness for real-world deployment.
\end{itemize}

\begin{table}[h!]
\caption{Agent evaluation on dynamic tariffs.}
\begin{center}
\begin{tabular}{|c|c|c|}
\hline
\textbf{cluster}&\multicolumn{2}{|c|}{\textbf{Evaluation results}} \\
\cline{2-3} 
\textbf{group} & \textbf{\textit{Financial cost}}& \textbf{\textit{Environmental Cost}}\\
\hline
1 & {1.005}$ \pm 0.1$ & {0.93}$\pm 0.2$ \\
2 &{ 1.03} $\pm 0.1$ & {0.84} $\pm 0.4$\\
3 & 1.01 $\pm 0.1$ & {0.88} $\pm 0.015$\\
\hline
\end{tabular}
\label{tab:results-pricing}
\end{center}
\end{table}

\subsection{Discussion}
The integrated framework described in this study successfully addresses the challenges that are key in managing diverse building energy consumption profiles and adapting to dynamic operational conditions. By combining clustering, forecasting, and constrained reinforcement learning, our approach demonstrates enhanced scalability, adaptability, and safety applied to building energy management.

The central component of this work is the clustering module, which identifies latent consumption patterns among buildings. This step not only reduces the complexity of dealing with numerous buildings with a variety of consumption profiles but also enables policy generalization and transfer learning by reducing the size of the problem to a small number of instances. The formation of clusters ensures that a single policy can be effectively deployed across multiple buildings exhibiting similar consumption characteristics, thus simplifying policy development and implementation. Our results show that even with limited observation windows—just one week of data—the clustering maintains consistency with year-long analyses, allowing for rapid classification and the immediate application of pre-learned policies. This capability underscores the method’s scalability and practical relevance: new buildings can be quickly integrated into the system, benefiting from previously trained policies without costly and time-consuming retraining processes.

The integration of an LSTM-based forecasting module further enhances the decision-making capabilities of the RL agent. By providing accurate short-term forecasts of energy prices, solar generation, and non-shiftable loads, the agent can anticipate changes and adjust its behavior in advance. This forecasting-driven lookahead is key for adapting to the non-stationarity commonly found in real-world energy environments, where demand, generation, and pricing fluctuate due to factors such as weather patterns, occupant behavior, and market dynamics. Improved predictive accuracy not only leads to better operational decisions but also mitigates risk, reduces energy costs, and strengthens the resilience of the entire system.

Moreover, the action masking mechanism ensures safe and domain-consistent reinforcement learning. By embedding prior knowledge of system constraints directly into the RL algorithm, we guarantee that the agent’s exploration remains confined to physically and operationally feasible actions. This approach reduces the likelihood of equipment degradation and network destabilization, thereby increasing the reliability and robustness of the system. The safety-focused design element also contributes to more efficient learning, as the agent’s policy gradient updates are derived solely from relevant and viable trajectories.

The combination of clustering, forecasting and constrained RL is the basis of the method's adaptability. The successful performance under stochastic tariff scenarios shows that it is capable to adapt its decision making process. In these experiments, pricing and environmental cost signals were perturbed to emulate real-world uncertainties in energy markets. Despite these perturbations, the trained agent maintained stable and cost-effective operation without retraining. This highlights the method’s transferability: not only can learned policies be applied to newly encountered buildings, but they can also be readily adapted to evolving market conditions by adjusting the masking constraints as needed. Thus, the method can evolve alongside changing policies, regulatory frameworks, and resource availabilities.

However, certain limitations and considerations remain. While the chosen  clustering techniques, and LSTM architectures proved effective, there is room to explore alternative more flexible clustering methods, or advanced forecasting models (e.g., Transformers) for enhanced performance. Future work may also consider the integration of the prediction model into the learning agent itself, allowing the RL policy to directly leverage anticipatory insights rather than relying solely on external inputs. Furthermore, the clustering-based policy assignment, while effective for known building types, introduces challenges when encountering buildings whose profiles do not align with existing clusters. In such cases, training a new policy from scratch becomes necessary, increasing operational complexity. Similarly, if a building’s consumption behavior changes significantly over time, the current framework lacks a dedicated mechanism to adapt its assigned policy. Addressing these issues may involve developing dynamic clustering methods, incremental policy adaptation modules, or online learning approaches to ensure robust and continuous optimality in evolving environments such as hierarchical RL. Additionally, extending the framework to incorporate other energy services—such as electric vehicle charging or interaction with distributed generation could reveal new challenges for the optimization framework. Finally, evaluating the approach under more extreme market scenarios, regulatory shifts, or unprecedented climatic conditions would further validate its overall robustness and applicability.

\section{Conclusion} \label{section:conclusion}
In this work, we presented a novel integrated framework for building energy management that combines clustering-based generalization, predictive modeling via LSTM, and safe reinforcement learning control. Our approach addresses critical challenges in modern energy systems, including the growing complexity and diversity of building consumption patterns, the non-stationarity of real-world conditions, and the necessity to ensure safety and feasibility in decision-making.

By leveraging hierarchical clustering on historical non-shiftable load data, we captured intrinsic consumption behaviors and grouped buildings into well-defined clusters. This clustering step enabled three key benefits: (1) policy generalization, by allowing a single policy to serve multiple buildings with similar consumption patterns; (2) transfer learning, by easily assigning pretrained policies to new buildings that fit known consumption profiles; and (3) targeted optimization, by applying cluster-specific strategies that handle each group’s unique operational challenges. Although effective, this method also introduces a limitation: if a building’s consumption profile does not fit any existing cluster, a new policy must be trained from scratch. Furthermore, if a building’s behavior significantly changes over time, our current framework lacks a module to dynamically reassign or adapt the policy, underscoring an avenue for future improvements.

The integration of an LSTM-based forecasting module further enhanced the agent’s decision-making by providing foresight into upcoming conditions. Forecast accuracy translated into improved adaptability to non-stationary environments, enabling proactive adjustments in response to shifting energy demands, renewable generation volatility, and fluctuating tariff structures. We highlight the potential of incorporating the prediction model directly into the reinforcement learning agent, a step that could streamline decision-making and potentially yield even more efficient energy management policies.

Our method’s focus on safe action selection, enforced through domain-inspired constraints and action masking, ensured the agent’s exploration remained confined to operationally and physically feasible actions. This design not only improved policy convergence and stability but also made the solution more practical for real-world deployment, where equipment longevity and grid stability are paramount.

The experiments conducted in the CityLearn environment demonstrated the validity and robustness of our approach. We achieved cost reductions, stable environmental impacts, and adaptability to dynamic market conditions without the need for full retraining. The ability to fine-tune constraints post-training and seamlessly adapt existing policies to changing conditions highlights the framework’s scalability and potential cost-effectiveness.

Yet, several directions remain to be explored. Future work could investigate more flexible clustering schemes or incremental learning methods that handle emerging building profiles and behavioral shifts without retraining from scratch. Incorporating advanced models, such as Transformers or graph neural networks, could improve forecasting accuracy further. Extending the approach to manage other energy services, such as electric vehicle charging or distributed generation coordination, could open new optimization opportunities. Finally, testing under extreme scenarios, regulatory changes, and unpredictable climatic events will be essential to fully validate the approach’s broad applicability.

In conclusion, this study provides a comprehensive, adaptable, and safe framework for building energy management. By uniting clustering, forecasting, and constrained RL, we have created a scalable solution that can efficiently manage the complexities and uncertainties of modern energy systems, laying a foundation for more intelligent, sustainable, and cost-effective building energy management practices.





\bibliographystyle{elsarticle-num} 
\bibliography{biblio}





\end{document}